\documentclass[journal,twoside]{IEEEtran}
\usepackage{microtype}
\usepackage{hyperref}
\usepackage{amsmath}
\usepackage{amsfonts}
\usepackage{algorithmic}
\usepackage{textcomp}
\usepackage{siunitx}
\usepackage{chemformula}
\usepackage{float}
\usepackage{bm}
\usepackage{booktabs}
\usepackage{gensymb}
\usepackage[capitalize]{cleveref}
\usepackage{cite}

\makeatletter
\let\MYcaption\@makecaption
\makeatother

\usepackage[font=footnotesize]{subcaption}

\makeatletter
\let\@makecaption\MYcaption
\makeatother

\def\Lair{L_{\rm air}}
\def\Lobj{L_{\rm obj}}
\def\Lobs{L_{\rm obs}}
\def\Lref{L_{\rm ref}}
\def\Lamb{L_{\rm amb}}
\def\LD{L_{\rm D}}
\def\LG{L_{\rm G}}
\def\Tair{T_{\rm air}}
\def\Thatair{\widehat{T}_{\rm air}}
\def\emiss{\varepsilon}
\def\bfemiss{\bm{\varepsilon}}
\def\lambdaS{\lambda_{\rm sat}}
\def\Ldata{\mathcal{L}_{\rm data}}
\def\Lreg{\mathcal{L}_{\rm reg}}
\def\yhat{\widehat{y}}
\def\dhatN{\widehat{d}_{\rm N}} 
\def\dhatA{\widehat{d}_{\rm A}} 
\def\dhatD{\widehat{d}_{\rm D}} 
\def\dmax{d_{\rm max}}
\def\ratioA{\gamma_{\rm A}}
\def\ratioD{\gamma_{\rm D}}

\newcommand{\Frac}[2]{{{#1}/{#2}}}  
\newcommand{\figref}[1]{(\subref{#1})} 

\DeclareMathOperator*{\argmin}{\arg\!\min}

\begin{document}

\title{Ozone Cues Mitigate Reflected Downwelling Radiance in LWIR Absorption-Based Ranging}

\author{Unay Dorken Gallastegi,
Wentao Shangguan,
Vaibhav Choudhary,
Akshay Agarwal,
\\
Hoover Rueda-Chac{\'o}n,
Martin J. Stevens,
and
Vivek K Goyal
\thanks{U. Dorken Gallastegi, W. Shangguan, V. Choudhary and V. K. Goyal are with the Department of Electrical and Computer Engineering, Boston University, Boston, MA 02215 USA
        (e-mail: \{udorken,wentaos,vchoudh\}@bu,edu; v.goyal@ieee.org).}
\thanks{A. Agarwal was with Boston University at the initiation of this work
        (e-mail: akshayagarwal019@gmail.com).}
\thanks{H. Rueda-Chac{\'o}n is with the Department of Computer Science, Universidad Industrial de Santander, Bucaramanga, 680002, Colombia
        (e-mail: hfarueda@uis.edu.co).}
\thanks{M. J. Stevens is with the National Institute of Standards and Technology, Boulder, CO, 80305 USA
        (e-mail: martin.stevens@nist.gov).}
\thanks{This work was supported
in part by the US Defense Advanced Research Projects Agency (DARPA)
Invisible Headlights program under contract number HR0011-20-S-0045,
in part by the US National Science Foundation under grants 1955219 and 2437371,
in part by a gift from Dr.\ John Z. Sun,
and
in part by a 2024 Guggenheim Fellowship.}
}

\maketitle

\begin{abstract} 
Passive long-wave infrared (LWIR) absorption-based ranging relies on atmospheric absorption to estimate distances to objects from their emitted thermal radiation. First demonstrated decades ago for objects much hotter than the air and recently extended to scenes with low temperature variations, this ranging has depended on reflected radiance being negligible. Downwelling radiance is especially problematic, sometimes causing large inaccuracies. In two new ranging methods, we use characteristic features from ozone absorption to estimate the contribution of reflected downwelling radiance.
The quadspectral method gives a simple closed-form range estimate from four narrowband measurements,
two
at a water vapor absorption line and two at an ozone absorption line.
The hyperspectral method uses a broader spectral range to improve accuracy while also providing estimates of temperature, emissivity profiles, and contributions of downwelling from a collection of zenith angles. Experimental results demonstrate improved ranging accuracy, in one case reducing error from over 100\,m when reflected light is not modeled to 6.8\,m with the quadspectral method and 1.2\,m with the hyperspectral method.
\end{abstract}


\section{Introduction}
\label{sec:introduction}

Range imaging is essential to a wide variety of fields, including agriculture, mapping, navigation, and security~\cite{LiuPRH:2020}.
Obtaining accurate range measurements in the dark is a significant challenge, as conventional systems often rely on visible light.
This becomes especially problematic when active methods, such as lasers or headlights, are not desirable due to stealth or power constraints.
In such settings, ambient infrared radiation emitted by objects in the scene can be used to estimate range passively, without requiring external illumination.
Though stereo disparity, a common passive method, can be applied to thermal imagery, its reliance on spatial texture makes it unreliable at long distances or for scenes with little texture~\cite{matthies1994stochastic,sibley2007bias,Clark2010,Gao2020}---an issue frequently encountered in thermal images~\cite{Gurton:14,bao2023heat,bao2024thermal}.
Absorption-based ranging offers a promising alternative by harnessing the wavelength dependence of atmospheric absorption~\cite{leonpacher1983passive, hawks2006passive, anderson2010monocular, anderson2011flight, vincent2011passive, hawks2013short, yu2017passive, yu2019threechannel, Nagase2022, 10877411, Kaariainen2024, Kushida2024}.
Using models for how ambient thermal radiation is spectrally attenuated by the atmosphere, this method enables passive depth estimation both day and night, without relying on scene texture or active illumination.
Exploiting absorption in this way is relatively easy for objects much hotter than their surroundings~\cite{leonpacher1983passive, hawks2006passive, hawks2013short, vincent2011passive, anderson2010monocular, yu2017passive, yu2019threechannel, anderson2011flight, Nagase2022, Kushida2024}.
In natural scenes, however, absorption-based ranging relies not only on sufficient radiance reaching the sensor, but also on a temperature contrast between the object and the surrounding air~\cite{cermak2017eleven}.
A low temperature contrast reduces the strength of the distance-dependent absorption features, limiting ranging performance~\cite{gallastegi2024absorption}.
While the generally poor conditioning associated with low temperature contrast can be counteracted by using many spectral measurements, another fundamental difficulty remains: 
the thermal emission signals of interest are often comparable in magnitude to the reflected thermal radiance—a phenomenon commonly referred to as \emph{ghosting}~\cite{Gurton:14,bao2023heat,bao2024thermal}.

Complications due to reflection are present in various 3D imaging methods and even in human depth perception~\cite{Higashiyama2012}.
In lidar, specular reflections fail to reach the detector unless a surface happens to be frontoparallel,
resulting in gaps in estimated range images or point clouds~\cite{Yang2011, Zhao2020},
while diffuse reflections can either enable multi-bounce non-line-of-sight imaging when intentionally modeled~\cite{FaccioVW:20} or introduce noise when not accounted for.
In stereo vision, specular reflections produce anomalous bright pixels that disrupt disparity estimation~\cite{BhatN:98},
whereas diffuse reflections shape the perceived brightness that underlies the entire method.
In structured light systems, reflections can create virtual objects or distort projected patterns, leading to incorrect depth estimation and scene misinterpretations~\cite{Whelan2018}.
While those issues have been previously addressed in the literature,
reflections are especially insidious in absorption-based ranging.
Distance inferences are based on wavelength-dependent features that are characteristic of light propagation over a distance,
so the reflection of light that already contains these features upon incidence at an object of interest will mislead absorption-based ranging methods~\cite{10877411}.
Since the amount of reflected light and the sources of reflected light are unknown, there are no obvious compensations.
This paper addresses inaccuracy due to reflected light
for the first time by using information rooted in ozone absorption to approximately separate emitted and reflected radiance.

Reflected light that has traveled a long distance before reaching a field-of-view object of interest is the most problematic.
In particular, the \emph{downwelling radiance}---which refers to the thermal radiation from the overhead sky---travels long distances through the atmosphere before reaching the ground, accumulating similar water vapor absorption features as light traveling far horizontally through the atmosphere.
For reflective materials, the downwelling radiance can significantly affect measurements and thus disrupt inferences made from hyperspectral measurements~\cite{Manolakis2019}.
Methods to compensate for reflected downwelling in temperature and emissivity separation have generally assumed a single transmittance function for entire scenes~\cite{borel2008error, borel2011recent, 10.1117/12.2239138,adler2014long};
contrarily, absorption-based ranging requires spatially varying transmittance as the fundamental basis for distance estimation.

\begin{figure}
    \begin{subfigure}[t]{0.32\linewidth}
        \centering
        \includegraphics[width=\linewidth]{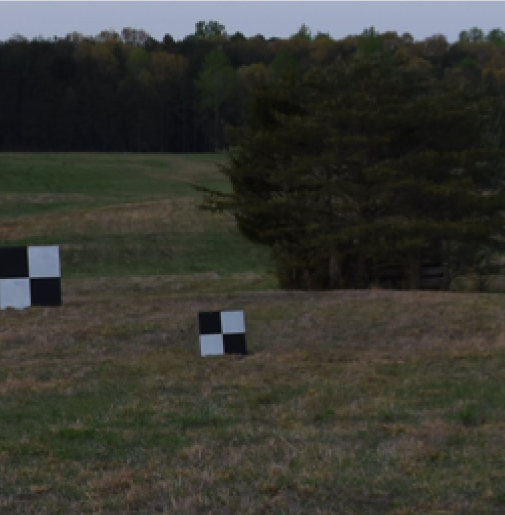}
        \caption{RGB photograph \\ (reference)}
        \label{fig:intro_RGB}
    \end{subfigure}
    \begin{subfigure}[t]{0.32\linewidth}
        \centering
        \includegraphics[width=\linewidth]{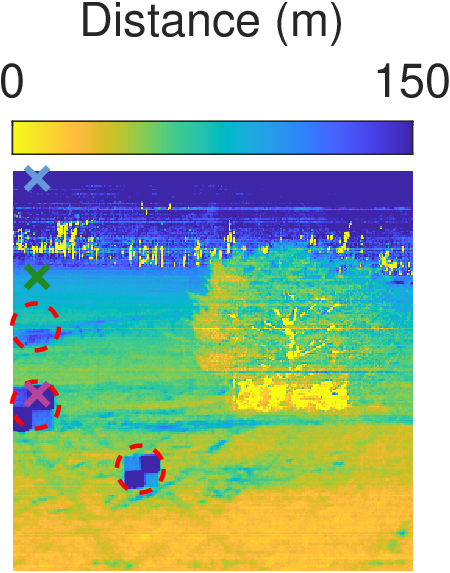}
        \caption{Neglecting \\ downwelling radiance}
        \label{fig:intro_no_downwelling}
    \end{subfigure}
    \begin{subfigure}[t]{0.32\linewidth}
        \centering
        \includegraphics[width=\linewidth]{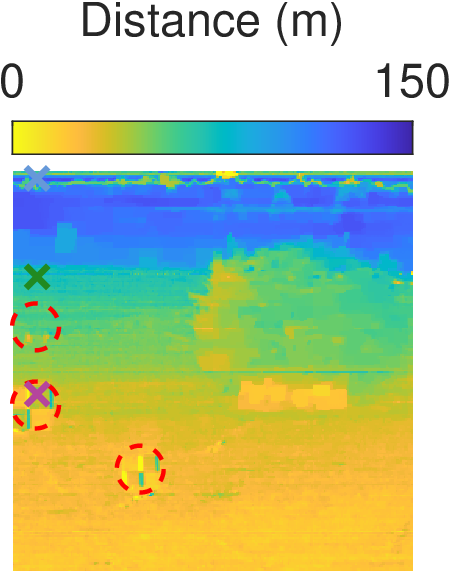}
        \caption{Accounting for \\ downwelling radiance}
        \label{fig:intro_with_downwelling}
    \end{subfigure}
    \\
    \begin{subfigure}[b]{\linewidth}
        \centering
        \includegraphics[width=\linewidth]{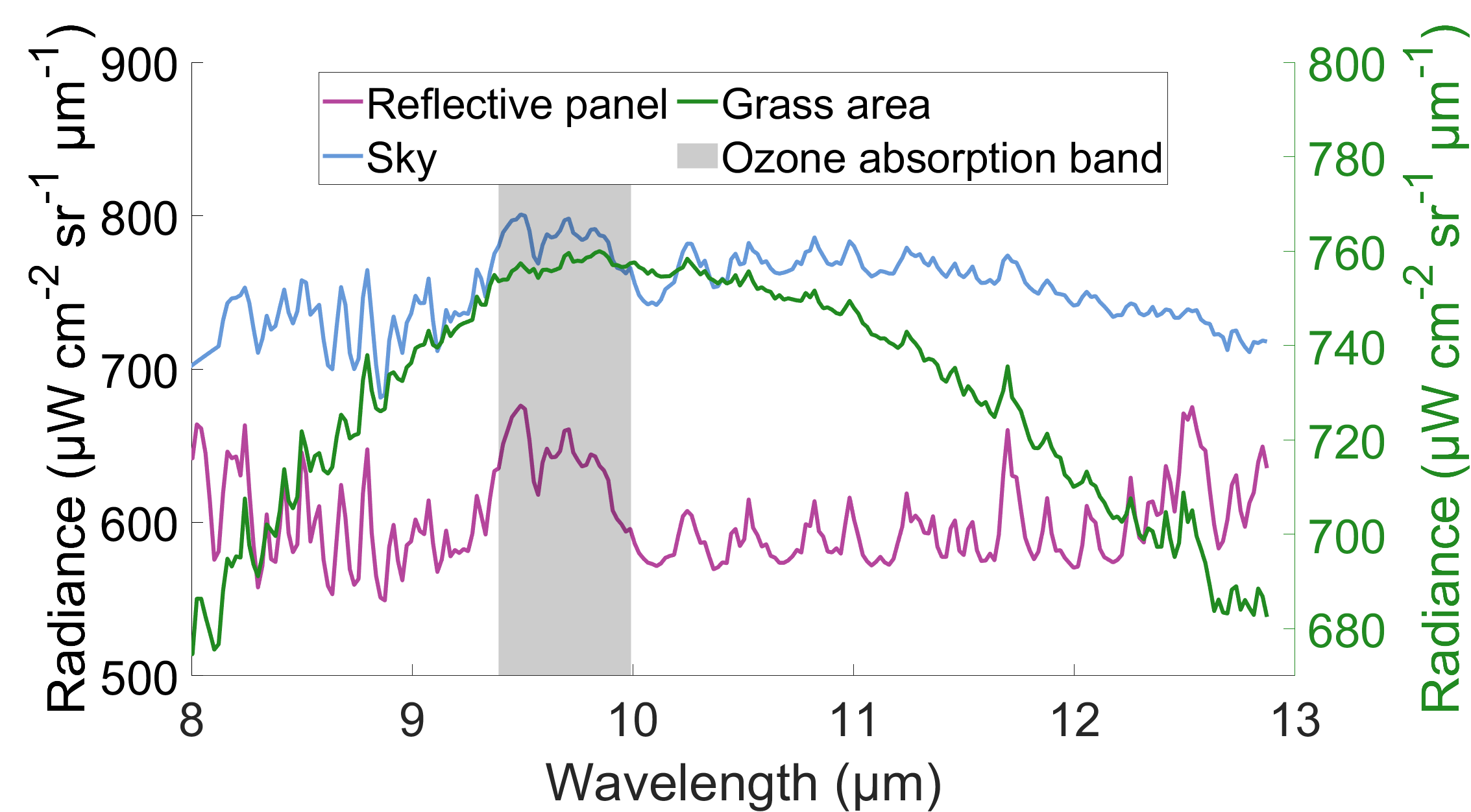} 
        \caption{Ozone absorption band (black curve) and measured radiances at three different locations of the scene: reflective panel (magenta curve), grass area (green curve), sky (blue curve)}
        \label{fig:intro_measurements}
    \end{subfigure}
    \caption{
    Effect of downwelling radiance in absorption-based ranging:
    \figref{fig:intro_RGB} RGB photograph of the scene shown for reference but not used in any computations;
    \figref{fig:intro_no_downwelling} hyperspectral absorption-based ranging results when neglecting downwelling radiance;
    \figref{fig:intro_with_downwelling} hyperspectral absorption-based ranging results when accounting for downwelling radiance (discussed in Section~\ref{sec:Hyperspectral_estimation});
    \figref{fig:intro_measurements} hyperspectral measurements at reflective panel, grass area, and sky pixels
    (to improve clarity, the green curve uses a distinct vertical scale provided on the right side).
    Pixels plotted in \figref{fig:intro_measurements} are shown with $\times$'s in \figref{fig:intro_no_downwelling} and \figref{fig:intro_with_downwelling}.
    The ozone absorption band is shown in gray.
    In natural scenes, where temperature variations are minimal, reflection from the overhead sky can contribute significantly to the measurements and cause range overestimations depending on the reflectivity of the material.
    The central innovation of this paper is to use measurements in the ozone absorption band to infer the strengths of downwelling radiance contributions from a set of zenith angles.
    This helps to separate the reflected downwelling radiance and improve the absorption-based ranging.
    }
    \label{fig:intro}
\end{figure}

\Cref{fig:intro} illustrates the effect of downwelling radiance in passive thermal absorption-based ranging.
For the scene shown in the RGB photograph in \cref{fig:intro_RGB},
when downwelling radiance is neglected,
the ranges of
reflective surfaces in the scene are overestimated,
as shown with red circles in \cref{fig:intro_no_downwelling}.
\Cref{fig:intro_measurements} compares spectral measurements from an overestimated region
(\emph{Reflective panel})
and an accurately estimated region (\emph{Grass area}) against a sky pixel.
Pixel locations are marked with $\times$'s on the depth maps (\cref{fig:intro_no_downwelling,fig:intro_with_downwelling}).
The sharp variations in these spectra arise from atmospheric absorption and emission.
Grass, which is known to behave as a near blackbody (non-reflective)~\cite{Manolakis2019}, exhibits spectral variations primarily due to absorption in the atmosphere between the object and the sensor.
In contrast, the variations in the spectrum at the reflective panel closely match the variations in the sky spectrum, suggesting that range overestimation may be due to amplified absorption features from downwelling radiance.

Most of the long-wave infrared (LWIR) band (8~\si{\micro\meter}--13~\si{\micro\meter}) is influenced by water vapor, which is present at all levels of the atmosphere and therefore contributes to both ground-level absorption and downwelling radiance.
However, the ozone absorption feature at 
9.5\,\si{\micro\meter} provides a way to distinguish between these two effects. 
Downwelling radiance exhibits a strong ozone absorption feature due to emissions from the ozone layer~\cite{10.1117/12.2239138}, whereas ozone absorption at ground level is negligible.
This unique spectral signature of ozone in downwelling radiance provides a key indicator of reflected downwelling contributions in the measurements and can be leveraged to correct overestimated range estimates, as previewed in \cref{fig:intro_with_downwelling}.

In this paper, we introduce two novel methods to mitigate errors caused by reflected downwelling radiance.
The first approach, \emph{quadspectral estimation},
uses four wavelengths (two wavelengths near water vapor absorption and two wavelengths near ozone absorption), offering a closed-form solution that is computationally trivial.
The second approach, \emph{hyperspectral estimation},
leverages a large number of spectral measurements while modeling downwelling radiance spectra.
In addition to improving range estimation, the hyperspectral method jointly estimates the temperature and emissivity profile.
Furthermore, it has the potential to provide insights into other object properties, such as surface normals.

\section{Ranging Problem and Measurement Model}
\label{sec:measurement_model}

Absorption-based ranging exploits the atmosphere's absorption lines to estimate distances by inverting Beer's law. 
Variations in the atmospheric transmittance spectrum enable inversion of the forward model using at least two spectral measurements at neighboring wavelengths to exploit the smoothness of emissivity and Planck's law.
Many studies explore the use of intensity attenuation for 3D imaging, including both active and passive methods.
Active methods~\cite{asano2020depth, kuo2021non, kuo2021surface, tsiotsios2017near, fujimura2018photometric, Kaariainen2024}
can readily create strong absorption features in the measurements.
Studies on passive absorption-based ranging are concentrated on tracking hot objects such as jet engines, missiles, light bulbs, or objects around 100\si{\celsius}~\cite{leonpacher1983passive, hawks2006passive, hawks2013short, vincent2011passive, anderson2010monocular, yu2017passive, anderson2011flight, Nagase2022}, where the thermal emission from the objects dominate the measurements.
In natural scenes, the temperature variation is limited to just a few degrees~\cite{cermak2017eleven},
causing absorption features to be weak and thus making ranging harder~\cite{gallastegi2024absorption}.
Under these challenging conditions, experimental results show reliable ranging for highly emissive materials such as grass; however, for highly reflective materials, the contribution from reflected downwelling radiance (sky radiance) can significantly disturb the estimated range under the open sky~\cite{gallastegi2022absorption, 10877411}.

\begin{figure}
    \centering
    \includegraphics[width=\linewidth]{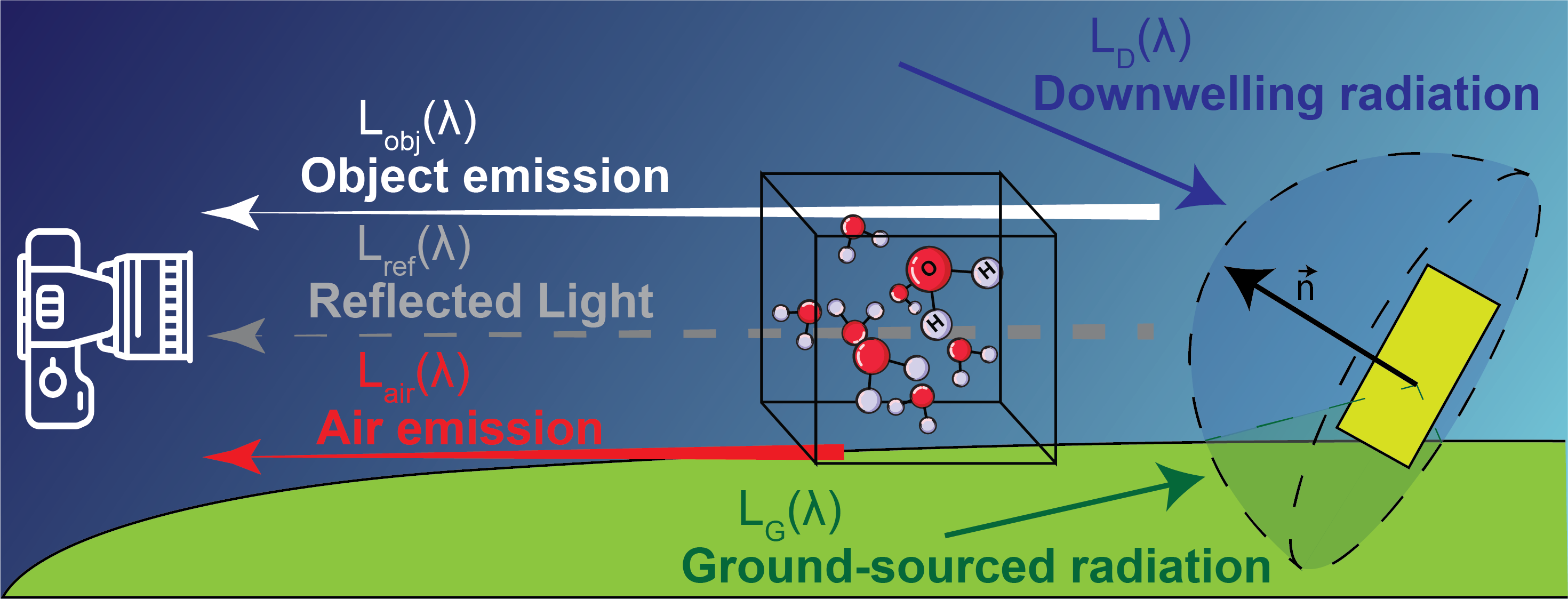}
    \caption{Conceptual figure for radiative transfer model. White and red arrows represent contributions from object emission and air emission to the observed spectrum, respectively.
    The gray arrow denotes reflected thermal radiation from incident ambient thermal radiation from all angles of the hemisphere, assuming a diffuse reflection model.
    The ambient thermal radiation is decomposed into downwelling and ground-sourced components.}
    \label{fig:measurement_model}
\end{figure}

We adopt the radiative transfer and instrument measurement models from~\cite{10877411}.
As illustrated in \cref{fig:measurement_model},
the radiance arriving at the sensor is a combination of three terms:
object emission $\Lobj(\lambda)$, reflected light $\Lref(\lambda)$, and air emission $\Lair(\lambda)$.
Each object in the scene emits thermal radiation.
Some of this reaches the sensor directly and some after reflection off other scene elements.
The total radiance from emission and reflection reaches the sensor after attenuation by the wavelength-dependent atmospheric transmission function, resulting in the first two terms.
The third term is thermal radiation from the atmosphere along the propagation path.
We now model each term more explicitly.

A blackbody at temperature $T$ emits thermal radiation following Planck's law of blackbody radiation,
\begin{equation}
\label{eq:Planck}
    B(\lambda; T) = \frac{2 h c^2}{\lambda^5} \frac{1}{e^{h c/ \lambda k_\mathrm{B} T} - 1},
\end{equation}
where $h$ is the Planck constant, $c$ is the speed of light, and $k_\mathrm{B}$ is the Boltzmann constant.
The radiance $B(\lambda; T)$ is typically expressed in units of microflicks (\si{\micro\watt \cdot sr^{-1} \cdot \centi\meter^{-2} \cdot \micro\meter^{-1}}).
Generally, objects are not blackbodies;
they emit only a portion of the blackbody radiation controlled by their wavelength-dependent emissivity $\emiss(\lambda)$.
The thermal radiation emitted from the object is 
\begin{equation}
\label{eq:Emitted_radiance}
\Lobj(\lambda) = \emiss(\lambda)B(\lambda; T).
\end{equation}
 
Alongside emitting thermal radiation, the object also reflects.
The reflectivity is the complement of the emissivity,  $1-\emiss(\lambda)$. 
We assume a Lambertian surface that diffusely reflects ambient thermal radiation.
Considering incoming ambient radiation $\Lamb(\lambda; \theta, \phi)$ from angles $(\theta,\phi) \in \Omega$ of the hemisphere,
the reflected radiation can be written as
\begin{align}
&\Lref(\lambda) = \frac{1 - \emiss(\lambda)}{\pi} \int_\Omega \Lamb(\lambda; \theta, \phi)\cos(\theta) \,d\omega \nonumber\\ 
&\hspace{2.8mm} = \frac{1 - \emiss(\lambda)}{\pi} \!\! \int_{0}^{2 \pi} \!\!\!\int_{0}^{\pi/2} \!\!\!\Lamb(\lambda; \theta, \phi) \cos(\theta) \sin(\theta) \, d\theta \, d\phi.
\label{eq:Reflected_radiance}
\end{align}

The combination of emission~\eqref{eq:Emitted_radiance} and reflection~\eqref{eq:Reflected_radiance} forms the object-leaving radiance.
The object-leaving radiance is attenuated by the atmosphere between the object and the sensor.
The portion that reaches the sensor is modeled with transmittance $\tau(\lambda)$.
Following Beer's law, assuming uniform concentration, the transmittance and distance are related as
\begin{equation}
\label{eq:Trasnmittance}
    \tau(\lambda; d) = 10^{-\Frac{\alpha(\lambda) d}{10}},
\end{equation} 
where $\alpha(\lambda)$ is the wavelength-dependent attenuation coefficient, in units of \si{\dB/\meter},
and $d$ is the range of the object in units of \si{\meter}.
In LWIR, scattering effects are not prominent,
and only absorption is modeled to represent the transmittance~\cite{minnaert1995light,manolakis2016hyperspectral}.
For extreme weather conditions or shorter wavelengths (approximately less than 1\,\si{\micro\meter} for air molecules), scattering can also be included~\cite{dubuc2021design}.

The atmosphere also emits radiation.
Following Kirchoff's law of thermal radiation, assuming thermal equilibrium and constant air temperature through the field of view of the sensor, the atmospheric emission is
\begin{equation}
\label{eq:Air_radiance}
\Lair(\lambda) = (1-\tau(\lambda;d))B(\lambda, \Tair).
\end{equation}
For remote sensing from satellites, the contribution from the atmosphere is often referred to as upwelling radiance.
It is more complex than the ground-level sensing that we consider in this paper, requiring an integral form because of gas content and temperature variations at different altitudes~\cite{manolakis2016hyperspectral}.

Combining the transmitted object-leaving radiance and atmospheric emission gives
the total radiance at the sensor as illustrated in \cref{fig:measurement_model}:
\begin{align}
\label{eq:radiance_model}
\Lobs(\lambda)
  & = \tau(\lambda;d)(\Lobj(\lambda) + \Lref(\lambda)) + \Lair(\lambda) \\
  & = 10^{-\Frac{\alpha(\lambda) d}{10}} (\emiss(\lambda)B(\lambda; T) \nonumber \\
  & \quad + \Lref(\lambda) - B(\lambda, \Tair)) + B(\lambda, \Tair),
\label{eq:radiance_model_alt}
\end{align}
where the second form uses substitution of
\eqref{eq:Emitted_radiance},
\eqref{eq:Trasnmittance},
and
\eqref{eq:Air_radiance}.
We aim to estimate $d$ from observations of this radiance, but this inverse problem is underdetermined.

Suppose we obtain measurements of
$\Lobs(\lambda)$
at $K$ wavelengths.
In \eqref{eq:radiance_model},
we could count as many as $4K$ unknowns,
with $K$ each from $\tau(\lambda)$, $\Lobj(\lambda)$, $\Lref(\lambda)$, and $\Lair(\lambda)$.
Assuming attenuation coefficients $\alpha(\lambda)$
and air temperature $\Tair$ are known,
the representation \eqref{eq:radiance_model_alt}
reduces to $2K+2$ unknowns:
$K$ each from
emissivity $\emiss(\lambda)$
and reflected radiation $\Lref(\lambda)$,
in addition to distance $d$ and object temperature $T$.
Methods that neglect
$\Lref(\lambda)$ and $\Lair(\lambda)$
can be accurate for very hot objects because
$\Lobj(\lambda)$ dominates the observed radiance~\cite{leonpacher1983passive, hawks2006passive, hawks2013short, vincent2011passive, anderson2010monocular, yu2017passive, yu2019threechannel, anderson2011flight, Nagase2022, Kushida2024}.
Recently introduced methods that include $\Lair(\lambda)$
are accurate without high object temperatures
so long as $\Lref(\lambda)$ remains negligible~\cite{10877411} 
as discussed in \cref{sec:ozone_absorption}. 
Our challenge here is to maintain accuracy even when $\Lref(\lambda)$ makes a significant contribution to the radiance at the sensor.
We seek mechanisms to estimate $\Lref(\lambda)$
to create feasibility for accurate estimation of $d$.

Reflected thermal radiation $\Lref(\lambda)$ originates from multiple environmental sources, arriving from various directions.  
The incoming ambient thermal radiation $\Lamb(\lambda; \theta, \phi)$ is decomposed as ground-sourced $\LG(\lambda; \theta, \phi)$ and downwelling $\LD(\lambda; \theta, \phi)$, shown in \cref{fig:measurement_model}:
\begin{equation}
\label{eq:ambient_decomposition}
\Lamb(\lambda; \theta, \phi) =  \LG(\lambda; \theta, \phi) + \LD(\lambda; \theta, \phi).
\end{equation}
Ground-sourced radiation $\LG(\lambda; \theta, \phi)$ consists of emissions from multiple objects with minimal atmospheric attenuation due to short propagation distances.  
In contrast, downwelling radiance $\LD(\lambda; \theta, \phi)$ travels long distances through the atmosphere before reaching the ground, undergoing significant absorption and emission, as illustrated in \cref{fig:intro_measurements}.  
Therefore the primary challenge is to distinguish and suppress downwelling radiance rather than ground-sourced reflections.
The ozone absorption band provides a key spectral feature to facilitate this distinction.

Our computational methods are based on measurements over some number of narrow bands or channels.
We approximated the discretized measurements by calculating the attenuation function $\alpha(\lambda)$
with a high-resolution spectral modeling software package
(SpectralCalc\footnote{Certain commercial products or company names are identified here to describe our study adequately.
Such identification is not intended to imply recommendation or endorsement by the National Institute of Standards and Technology, nor is it intended to imply that the products or names identified are necessarily the best available for the purpose.})~\cite{SpectralCalc}
that relies on a high-resolution transmission molecular absorption database (HITRAN2020)~\cite{GORDON2022107949}, assuming a Gaussian instrumental spectral response function (ISRF)~\cite{beirle2017parameterizing} with 40\,$\si{\nano\meter}$ full width at half maximum, consistent with the spectral bandwidth of the instrument used in the dataset.
The software uses a standard model of the atmosphere~\cite{coesa1976standard},
modified to match the temperature, pressure, and humidity values of the collected data.
These values were collected with a weather station on site.
We assume that the measurements are affected by additive white Gaussian noise.

\section{Ozone Absorption as a Distinguishing Feature}
\label{sec:ozone_absorption}

\begin{figure}
    \begin{subfigure}[b]{1\linewidth}
        \centering
        \includegraphics[width=\linewidth]{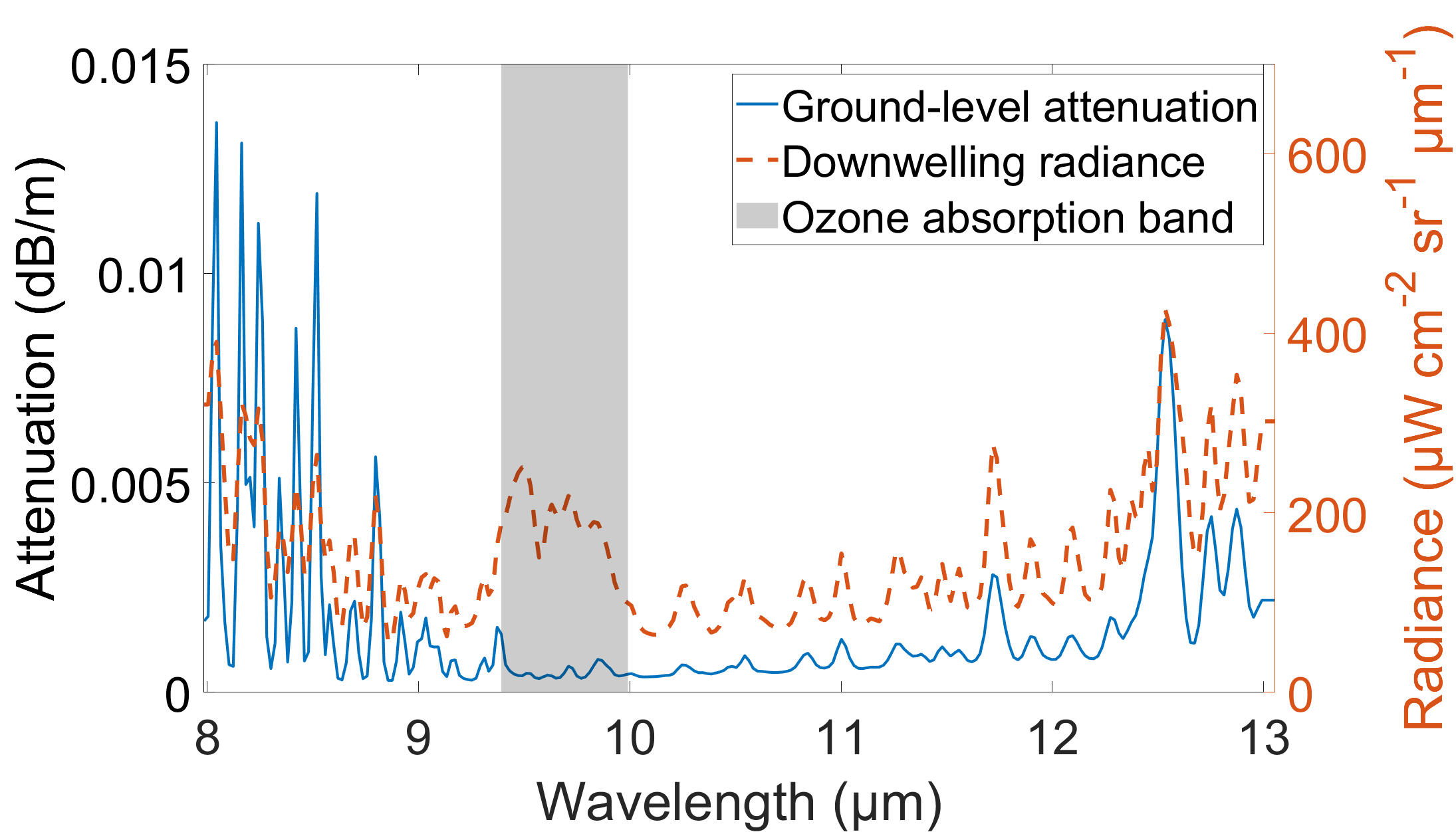}
        \caption{Ground-level attenuation (total across all molecular contributors)
        and downwelling radiance}
        \label{fig:ozone_feature_1}
    \end{subfigure}
    \begin{subfigure}[b]{1\linewidth}
        \centering
        \includegraphics[width=\linewidth]{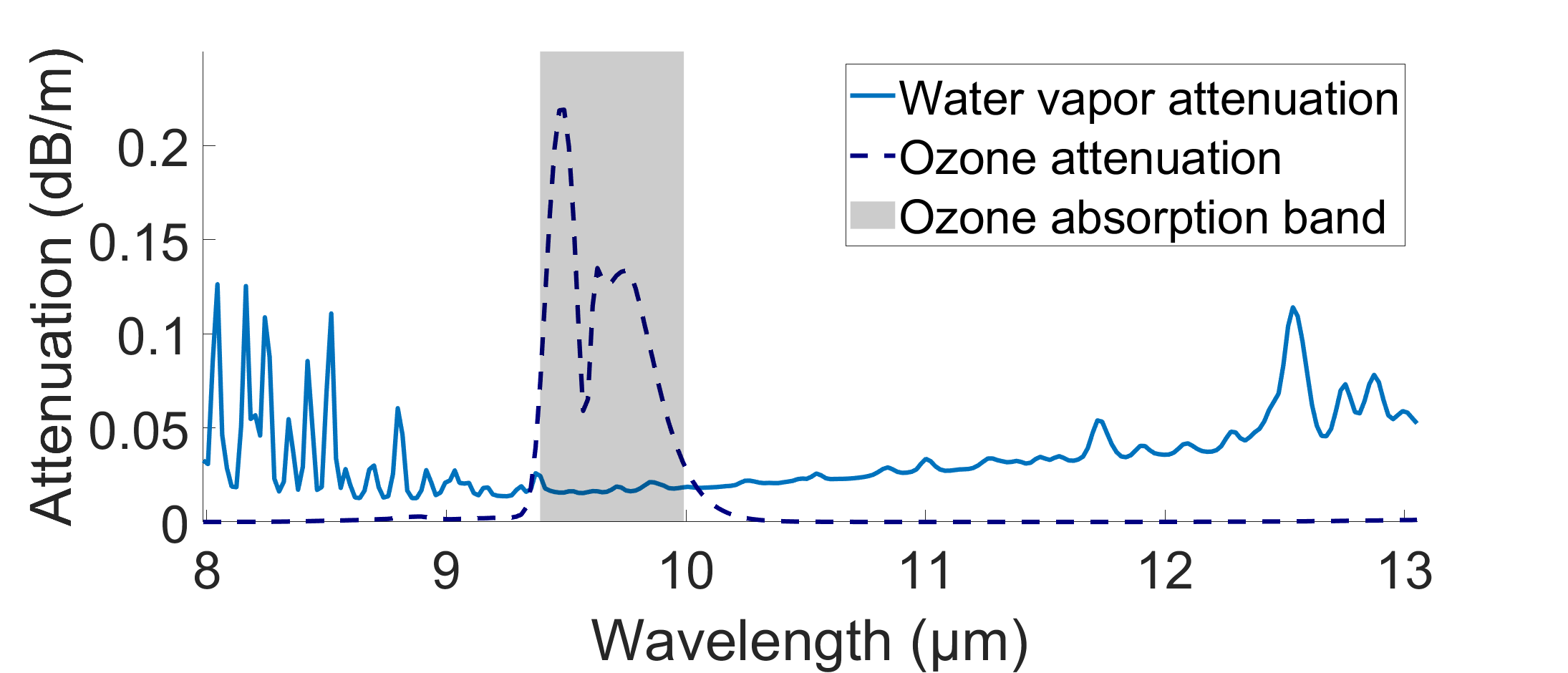}
        \caption{Ozone and water vapor attenuation}
        \label{fig:ozone_feature_2}
    \end{subfigure}
    \caption{
    Ozone absorption band in LWIR\@.
    \figref{fig:ozone_feature_1} Comparison of ground-level horizontal attenuation and vertical downwelling radiance computed using SpectralCalc~\cite{SpectralCalc} for the U.S. Standard Atmosphere~\cite{coesa1976standard}.
    \figref{fig:ozone_feature_2} Comparison of water vapor (0.1 volume mixing ratio) and ozone ($5\times10^{-5}$ volume mixing ratio) attenuation profiles.
    All spectra are calculated with 40\,\si{\nano\meter} Gaussian ISRF\@.
    Most of the absorption features at ground level are due to water vapor and are common to both ground-level attenuation and downwelling radiance.
    The absorption feature near 9.5~\si{\micro\meter} is due to ozone and is only observed in downwelling radiance.
    This is because ozone concentration at ground level is negligible, while the ozone layer at higher altitudes affects the downwelling radiance.
    }
    \label{fig:ozone_feature}
\end{figure}

A key challenge is to distinguish the reflected downwelling radiation and quantify the attenuation experienced by object emission. 
In LWIR, the downwelling radiance has a distinguishing ozone absorption feature at 9.5\,\si{\micro\meter} that is absent in ground-level transmittance.
This is due to the negligible concentration of ozone at ground level,
while the downwelling radiance originating from the atmosphere exhibits strong emission at this wavelength from the ozone layer.
Unless there is an unusual occurrence of an ozone plume along the horizontal propagation path (an unlikely event in natural scenes),
an observed ozone absorption feature is likely due to reflected downwelling radiance.
\Cref{fig:ozone_feature} illustrates the attenuation function at ground level, alongside the downwelling radiance perpendicular to the surface ($0^\circ$ zenith angle), modeled using the U.S. Standard Atmosphere~\cite{coesa1976standard}.
Most of the absorption lines are due to water vapor;
the absorption feature at 9.5\,\si{\micro\meter} corresponds to ozone and is unique to the downwelling radiance.

The strength of the ozone absorption feature can be measured by taking the difference of radiances at two nearby wavelengths.
The magnitude of the difference in the measurements gives a cue about the contribution of the reflected downwelling radiance; 
the higher the magnitude, the higher the contribution.
Based on the model in~\eqref{eq:radiance_model}, the contribution from downwelling radiance can vary from
factors such as the reflectivity (or emissivity) and surface normal of the object, and the solid angle looking towards the sky (quantifies the portion of the sky visible to the surface).

There are many ways the ozone absorption feature can be helpful. 
A threshold to the ozone absorption difference can be used to detect pixels that have major contribution from downwelling radiance, as done in~\cite{10877411}.
Naively discarding these pixels and running an inpainting algorithm could be beneficial.
Rather than throwing away all the information in these pixels,
however,
in the following sections
we exploit the correlation of the ozone cue to correct the overestimates.

\section{Quadspectral Estimation}
\label{sec:quadspectral_estimation}

Our quadspectral estimation is built upon bispectral estimation~\cite{10877411,asano2020depth,kuo2021non,Nagase2022,leonpacher1983passive,hawks2006passive},
which uses two wavelength measurements to provide a closed-form range estimate.
\Cref{sec:background_bispectral} provides a summary of the literature on bispectral estimation.  
Depending on the conditions, certain terms in the measurement model~\eqref{eq:radiance_model_alt} are neglected, leading to different variations of bispectral estimators.  
However, none of the existing estimators account for reflected thermal radiance.  
\Cref{sec:bispectral_downwelling} proposes a bispectral estimator that incorporates reflected thermal radiance by introducing an additional unknown term, $b$, representing the contribution of downwelling radiance.  
To estimate $b$, \Cref{sec:quadspectral_est_b} extends the approach by incorporating two additional wavelength measurements around the ozone absorption band, forming the quadspectral estimation method, where a total of four spectral bands are utilized.  
Finally, \Cref{sec:quadspectral_summary} summarizes the procedure for clarity and completeness.

\subsection{Background on Bispectral Estimation}
\label{sec:background_bispectral}

Bispectral estimators work under the assumption of Planck's function and the emissivity of solid materials both being smooth functions of wavelength.
The emissivity profiles of solid objects are smooth compared to the sharp transitions in the atmospheric transmittance \cite{Manolakis2019, borel2008error, borel2003artemiss}.%
\footnote{For illustrations of emissivity and atmospheric transmittance profiles, see, e.g.,~\cite[Fig.~7]{10877411}.}
Therefore for two nearby spectral bands $\lambda_1$ and $\lambda_2$ these values are approximately identical,
\begin{subequations}
\label{eq:bispectral_approximations}    
\begin{align}
    \emiss(\lambda_1) &\approx \emiss(\lambda_2).
\label{eq:emiss_same}
\end{align}
The smoothness of \eqref{eq:Planck} furthermore gives
\begin{align}
    B(\lambda_1; T)     &\approx B(\lambda_2; T), \\
    B(\lambda_1; \Tair) &\approx B(\lambda_2; \Tair).
\end{align}

\end{subequations}
The literature proposes several types of bispectral estimators
that treat \eqref{eq:bispectral_approximations} as holding with equality.
While there are variations
depending on which terms in~\eqref{eq:radiance_model_alt} are negligible, all previous works ignored the reflected radiance, assuming $\Lref(\lambda) \approx 0$~\cite{10877411,Nagase2022,leonpacher1983passive,hawks2006passive}.

For hot objects such as jet engines, missiles or light bulbs, the emitted radiance from the object is much larger than other thermal sources, such as the air
($\Lobj(\lambda) \gg \Lair(\lambda)$)
and the reflected radiance
($\Lobj(\lambda) \gg \Lref(\lambda)$).
Therefore, the contribution of air emission and reflected thermal radiance are ignored~\cite{Nagase2022,leonpacher1983passive, hawks2006passive} resulting in the approximate measurement model
\begin{align}
\Lobs(\lambda) &\approx \tau(\lambda;d)(\Lobj(\lambda))  \nonumber \\
               &=
               10^{-\Frac{\alpha(\lambda) d}{10}}
               (\emiss(\lambda)B(\lambda; T)).
\label{eq:radiance_model_hot}
\end{align}
Using \eqref{eq:radiance_model_hot},
if one divides observations at a pair of wavelengths
where \eqref{eq:bispectral_approximations} holds with equality,
the object emission factors cancel.
Taking the logarithm and rearranging gives
the bispectral estimate \emph{neglecting} air emission and reflected radiance:
\begin{equation}
    \dhatN = \frac{-10}{\alpha(\lambda_2) - \alpha(\lambda_1)}\log_{10}\!\left(\frac{\Lobs(\lambda_2)}{\Lobs(\lambda_1)} \right).
\label{eq:bispectral_hot_obj}
\end{equation}

When the object is around the same temperature as the air, the air emission should be accounted for in the measurement model~\cite{10877411}.
The approximate measurement model accounting for the air emission is
\begin{align}
\Lobs&(\lambda) \approx \tau(\lambda;d)(\Lobj(\lambda)) + \Lair(\lambda)  \nonumber \\
               &=
               10^{-\Frac{\alpha(\lambda) d}{10}}
               (\emiss(\lambda)B(\lambda; T) - B(\lambda, \Tair)) + B(\lambda, \Tair).
\label{eq:radiance_model_air}
\end{align}
The air temperature is assumed to be known, either from another sensor or estimated from a saturated absorptive band ($\lambdaS$), where the transmittance is close to zero ($\tau(\lambdaS) \approx 0$), via
\begin{equation}
    \Thatair = B^{-1}(\lambdaS; \Lobs(\lambdaS)),
\end{equation}
where $B^{-1}$ is the inverse of the blackbody function, otherwise known as brightness temperature~\cite{manolakis2016hyperspectral}.
Again assuming \eqref{eq:bispectral_approximations} holds with equality,
subtracting the now-known $B(\lambda, \Tair)$ from \eqref{eq:radiance_model_air} allows us to have a cancellation of
$\emiss(\lambda_i)B(\lambda_i; T) - B(\lambda_i, \Tair)$
in a ratio,
resulting in
the bispectral estimate including \emph{air emission}:
\begin{subequations}
\label{eq:bispectral_air}
\begin{equation}
    \dhatA = \frac{-10}{\alpha(\lambda_2) - \alpha(\lambda_1)}\log_{10} \ratioA,
\end{equation}
where
\begin{equation}
\label{eq:bispectral_air_ratio}
    \ratioA = \frac{\Lobs(\lambda_2) - B(\lambda_2;\Thatair)}{\Lobs(\lambda_1) - B(\lambda_1;\Thatair)}.
\end{equation}
\end{subequations}

\subsection{Bispectral Estimation Accounting for Reflected Thermal Radiance}
\label{sec:bispectral_downwelling}

When reflected radiance is not negligible,
the bispectral estimator accounting for
air emission $\dhatA$ in \eqref{eq:bispectral_air}
is inaccurate~\cite{10877411}.
We now derive an improvement.
Substituting \eqref{eq:radiance_model_alt} into \eqref{eq:bispectral_air_ratio},
assuming $B(\lambda,\Tair)$ is accurately canceled by $B(\lambda,\Thatair)$,
we get
\begin{equation}
    \ratioA = \frac{10^{-\Frac{\alpha(\lambda_2) d}{10}}(\emiss(\lambda_2)B(\lambda_2; T) + \Lref(\lambda_2) - B(\lambda_2, \Tair))}
                   {10^{-\Frac{\alpha(\lambda_1) d}{10}}(\emiss(\lambda_1)B(\lambda_1; T) + \Lref(\lambda_1) - B(\lambda_1, \Tair))}.
\end{equation}
Since we are not willing to assume $\Lref(\lambda_1) \approx \Lref(\lambda_2)$, the trailing factors in the numerator and denominator do not approximately cancel.
Instead,
we add and subtract $\Lref(\lambda_1)$ from the object emission factor in the numerator
to obtain
\begin{align}
    &\frac{\emiss(\lambda_2)B(\lambda_2; T) + \Lref(\lambda_2) - B(\lambda_2, \Tair)}
          {\emiss(\lambda_1)B(\lambda_1; T) + \Lref(\lambda_1) - B(\lambda_1, \Tair)} \nonumber \\
    &\quad =
        \frac{\emiss(\lambda_2) B(\lambda_2; T) + \Lref(\lambda_1) - B(\lambda_2, \Tair)}
             {\emiss(\lambda_1) B(\lambda_1; T) + \Lref(\lambda_1) - B(\lambda_1, \Tair)}
         \nonumber \\
    &\qquad +
        \frac{ \Lref(\lambda_2) - \Lref(\lambda_1) }
             { \emiss(\lambda_1) B(\lambda_1; T) + \Lref(\lambda_1) - B(\lambda_1, \Tair) },
\end{align}
where the first term is approximately 1 because of \eqref{eq:bispectral_approximations}.
This highlights that knowledge of the residual in the numerator,
\begin{equation}
\label{eq:b_definition}
    b = 10^{-\Frac{\alpha(\lambda_2) d}{10}}(\Lref(\lambda_2) - \Lref(\lambda_1)),
\end{equation}
would allow us to modify $\ratioA$ such that it reduces to a simple exponential in $d$.
Using this $b$, a bispectral estimate including \emph{downwelling} is obtained:
\begin{subequations}
    \label{eq:bispectral}
\begin{equation}
    \dhatD = \frac{-10}{\alpha(\lambda_2) - \alpha(\lambda_1)} \log_{10} \ratioD,
\end{equation}
where
\begin{equation}
    \ratioD = \frac{\Lobs(\lambda_2) - B(\lambda_2;\Tair) - b}
                   {\Lobs(\lambda_1) - B(\lambda_1;\Tair)}.
\end{equation}
\end{subequations}
We should note, however, that unlike
$\dhatN$ in \eqref{eq:bispectral_hot_obj} and
$\dhatA$ in \eqref{eq:bispectral_air},
this $\dhatD$ is not yet computable because $b$ is unknown.

\subsection{Use of Ozone Absorption Bands to Estimate $b$}
\label{sec:quadspectral_est_b}
We propose to use two more spectral measurements around the ozone absorption band to estimate $b$
for use in~\eqref{eq:bispectral}.
Due to the use of four total wavelengths, we call this method \emph{quadspectral estimation}.

In the decomposition \eqref{eq:ambient_decomposition},
the ground-sourced radiation is mainly a combination of different solid object emissions.
It is thus approximately identical $\LG(\lambda_1) \approx \LG(\lambda_2)$ at nearby wavelengths $\lambda_1$ and $\lambda_2$ from \eqref{eq:bispectral_approximations}. 
Therefore, the difference
$\Lref(\lambda_2) - \Lref(\lambda_1)$ in
\eqref{eq:b_definition}
cancels the ground-sourced radiances leaving
only a difference of downwelling radiances,
\begin{align}
    b &= 10^{-\Frac{\alpha(\lambda_2) d}{10}}(\Lref(\lambda_2) - \Lref(\lambda_1)) \nonumber \\
     &\approx \tau(\lambda_2)\frac{1 - \emiss(\lambda_2)}{\pi} \nonumber \\ 
     & \hspace{1cm} \int_{\Omega}(\LD(\lambda_2; \theta, \phi) - \LD(\lambda_1; \theta, \phi)) \cos(\theta) \, d\omega.
\end{align}

Downwelling radiances have been well-studied in modeling the temperature and gas concentration from the sky to the ground~\cite{coesa1976standard}.
The ozone absorption line is unique to downwelling radiance in our geometry;
ozone concentration at ground level is negligible,
so horizontal transmittance at the surface is nearly unaffected.
Therefore, for two wavelengths at the ozone absorption, $\lambda_3$ and $\lambda_4$, the transmittance terms are almost identical and close to unity $\tau(\lambda_3) \approx \tau(\lambda_4) \approx 1$.
For the attenuation function discussed in Section~\ref{sec:measurement_model}, the values at these wavelengths, computed using the U.S. Standard Atmosphere~\cite{coesa1976standard}, are nearly the same:
$\alpha(\lambda_3) = 3.98\times10^{-4}\,\si{dB/\meter}$
and
$\alpha(\lambda_4) = 3.31\times10^{-4}\,\si{dB/\meter}$.
Treating the transmittances as equal leads to approximating
the difference of observed measurements at the ozone absorption line as 
\begin{align}
&\Lobs(\lambda_4) - \Lobs(\lambda_3) \approx \nonumber \\
&\tau(\lambda_3)\frac{(1 - \emiss(\lambda_3))}{\pi}\int_{\Omega}(\LD(\lambda_4; \theta, \phi) - \LD(\lambda_3; \theta, \phi)) \cos(\theta) \, d\omega,
\label{eq:meas_diff_ozone}
\end{align}
where smooth object emissions cancel each other in the difference.

\begin{figure}
    \centering
    \begin{subfigure}[b]{0.48\linewidth}
        \centering
        \includegraphics[width=\linewidth]{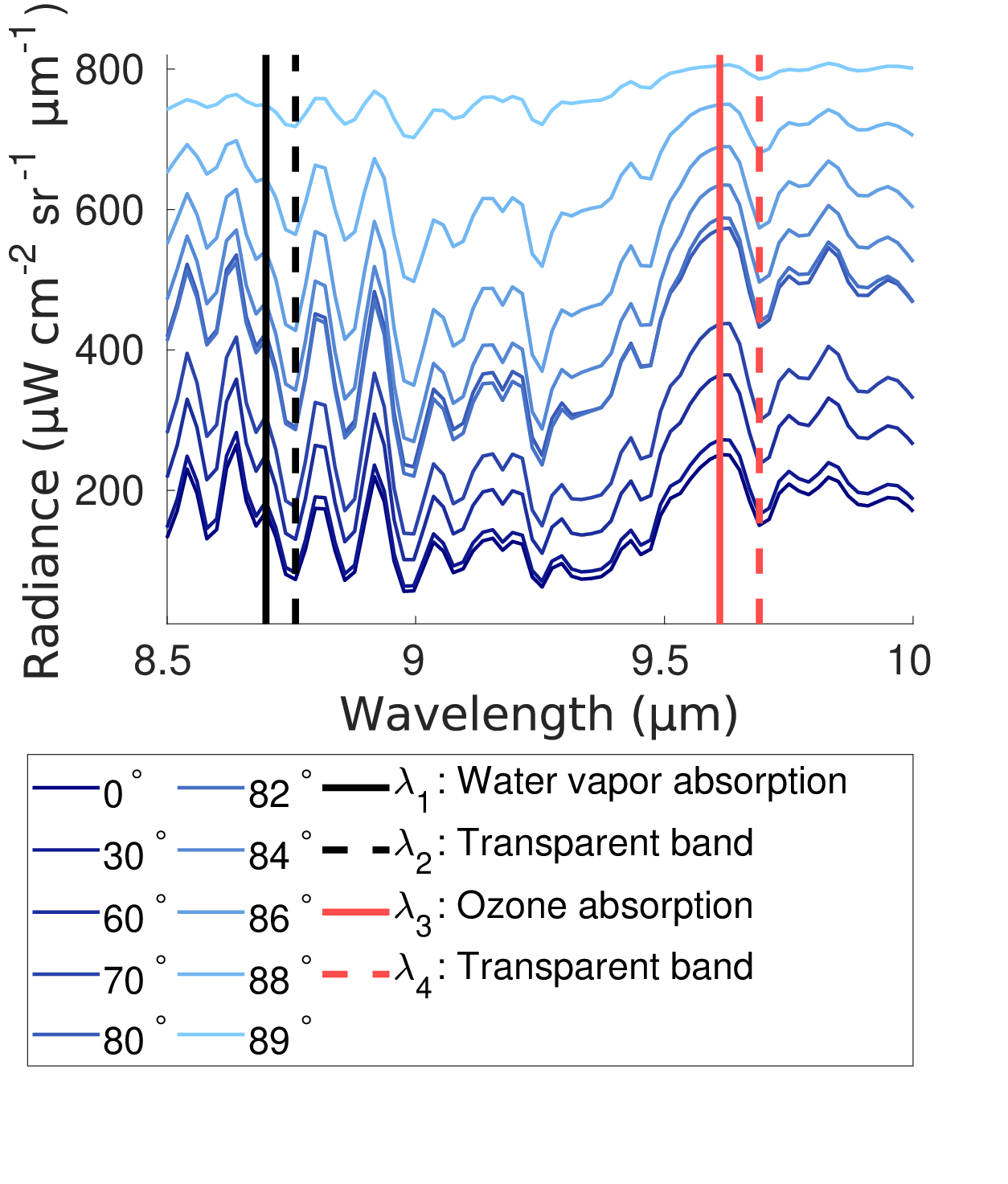}
        \caption{Simulated downwelling radiances \\ at different zenith angles.}
        \label{fig:Water_ozone_correlation_1}
    \end{subfigure}
    \hfill
    \begin{subfigure}[b]{0.48\linewidth}
        \centering
        \includegraphics[width=\linewidth]{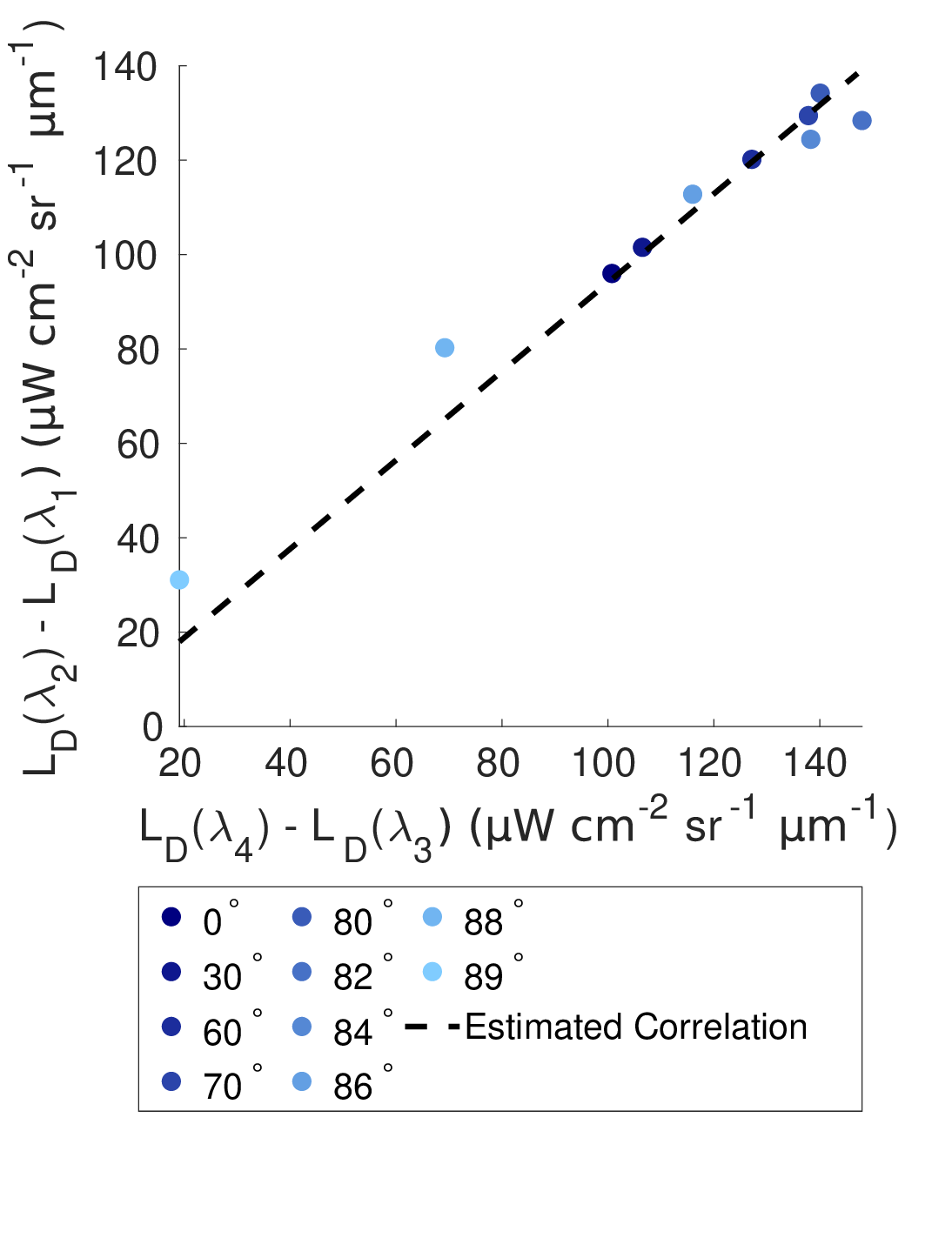}
        \caption{Correlation between the water \\ vapor and ozone features.}
        \label{fig:Water_ozone_correlation_2}
    \end{subfigure}
    \caption{Downwelling radiances at various zenith angles.
    \figref{fig:Water_ozone_correlation_1} Downwelling radiances calculated using SpectralCalc\cite{SpectralCalc} for 10 zenith angles ranging from $0^\circ$ to $90^\circ$, bottom to top.
    The ozone absorption band and one water vapor absorption band are highlighted with solid red and black vertical lines respectively.
    The dashed curves represent the nearby transparent bands to the absorption line with the same color encoding.
    \figref{fig:Water_ozone_correlation_2} Correlation between ozone difference (x-axis) and water vapor difference (y-axis), where blue circles represent the 10 zenith angles in \figref{fig:Water_ozone_correlation_1}.
    Shown water vapor features are at 8.58\,\si{\micro\meter}--8.64\,\si{\micro\meter} and the ozone features are at 9.49\,\si{\micro\meter}--9.57\,\si{\micro\meter}.
    The resulting correlation coefficient is
    $s = 0.94$,
    which is used to compute the estimate $\hat{b}$ of the bias $b$
    by scaling the measured radiance difference at the ozone absorption band, as described in~\eqref{eq:bias_estimate}.}
    \label{fig:Water_ozone_correlation}
\end{figure}

We assume a linear relationship between the water vapor difference and the ozone difference in the downwelling radiances, 
\begin{equation}
L_D(\lambda_2; \theta, \phi) - L_D(\lambda_1; \theta, \phi)
= s (L_D(\lambda_4; \theta, \phi) - L_D(\lambda_3; \theta, \phi)),
\label{eq:slope_downwelling}
\end{equation}
where $s$ is the slope of the linear relation.
\Cref{fig:Water_ozone_correlation_1} shows simulated downwelling radiances at
several zenith angles ranging from $0^\circ$ to $90^\circ$, and \cref{fig:Water_ozone_correlation_2} shows
that the differences indeed approximately follow the linear relation \eqref{eq:slope_downwelling}
when the water vapor feature at 8.58\,\si{\micro\meter}--8.64\,\si{\micro\meter}
is used.
The deviation from the linear relation depends on the choice of water vapor absorption pair, some wavelength selections being more correlated than others. 
Using the U.S. Standard Atmosphere~\cite{coesa1976standard} the slope $s$ is obtained by simulating downwelling radiances at multiple zenith angles in SpectralCalc~\cite{SpectralCalc} using an upward-looking path (observer at 0 km, target at 600 km, linelist HITRAN2020). SpectralCalc’s radiative-transfer model includes the ozone layer and other standard atmospheric absorbers through the selected atmospheric profile.

To estimate the term $b$ using the ozone correlation, we extend the gray-body assumption to other wavelengths:
$\emiss(\lambda_1) \approx \emiss(\lambda_2) \approx \emiss(\lambda_3) \approx \emiss(\lambda_4)$.
Furthermore, we assume that the ground level transmittances at $\lambda_2$, $\lambda_3$, and $\lambda_4$ are almost the same, $\tau(\lambda_2) \approx \tau(\lambda_3) \approx \tau(\lambda_4)$.
For the attenuation function discussed in \cref{sec:measurement_model}, the values at these wavelengths are nearly the same:
$\alpha(\lambda_2) = 2.98\times10^{-4}\,\si{dB/\meter}$,
$\alpha(\lambda_3) = 3.98\times10^{-4}\,\si{dB/\meter}$,
and
$\alpha(\lambda_4) = 3.31\times10^{-4}\,\si{dB/\meter}$.
With these assumptions $b$ is estimated as
\begin{align}
\label{eq:bias_estimate}
\hat{b} &= s (\Lobs(\lambda_4) - \Lobs(\lambda_3)) \nonumber\\
  &\approx \tau(\lambda_2)\frac{(1 - \emiss(\lambda_2))}{\pi} \nonumber \\
  & \hspace{1cm} \int_{\Omega}(\LD(\lambda_2; \theta, \phi) - \LD(\lambda_1; \theta, \phi)) \cos(\theta) \, d\omega,
\end{align}
where $s$ is computed from \eqref{eq:slope_downwelling}.
The estimate $\hat{b}$ represents a scaled version of the observed radiance difference at the ozone absorption band, as defined in \eqref{eq:meas_diff_ozone}.
After estimating $b$, the range considering reflected radiation can be calculated using \eqref{eq:bispectral}.

\subsection{Summary of Quadspectral Estimation}
\label{sec:quadspectral_summary}

In summary, we consider measurements at wavelengths $\lambda_1$, $\lambda_2$, $\lambda_3$, $\lambda_4$,
with ground-level transmittances that satisfy
$\tau(\lambda_1) < \tau(\lambda_2) \approx \tau(\lambda_3) \approx \tau(\lambda_4) \approx 1$.
The differences in transmittance are mainly due to water vapor absorption.
The wavelengths $\lambda_3$ and $\lambda_4$ are at the ozone absorption bands where approximately only the downwelling radiance is nonsmooth and varying.
We assume a gray body object (constant emissivity profile) through the four wavelengths.
This assumption is likely to hold when ozone and water vapor absorption lines are chosen close to each other considering the smoothness of the emissivity profiles.
The bias term is found from the ozone difference image and the correlation between the water vapor and ozone features of downwelling radiances.
Then, the estimator in~\eqref{eq:bispectral} can be used for ranging, accounting for the downwelling radiance.

\section{Hyperspectral Estimation}
\label{sec:Hyperspectral_estimation}

In contrast to quadspectral estimation, hyperspectral estimation considers many more wavelengths across the spectrum.
More wavelengths give more information and hence allow us to rely on fewer or weaker assumptions.
Unlike the quadspectral method, the hyperspectral method does not have a closed-form solution;
instead, it is a model-based inversion through minimization of a loss function using gradient descent.
The loss function is composed of three terms:
a term for the fit between the measurements and the measurement model in~\eqref{eq:radiance_model_alt},
a regularization term for smooth emissivity,
and
a term that regularizes by total variation (TV) of distance.
Using emissivity smoothness as the basis for the first regularization term is a weaker assumption than the equality assumptions underlying the quadspectral method.
The second regularization term is not required for good conditioning
but is instead used for denoising.

The reflected radiation from the sky is represented by $Q$ discretized incoming downwelling radiances at different zenith angles calculated from SpectralCalc~\cite{SpectralCalc},
\begin{equation}
    \label{eq:discretized_reflected_light}
    \Lref(\lambda) = \frac{1 - \emiss(\lambda)}{\pi} \sum_{q=1}^Q \Omega_q L_{D, q}(\lambda),
\end{equation}
where $\Omega_q$ represents the projected solid angle~\cite{Arecchi2007FieldGuide} of the $q$th downwelling radiance.
Discretized downwelling radiations are shown in~\cref{fig:Water_ozone_correlation_1}.
The sum of the projected solid angles is constrained to be less than $\pi$, and they are to be estimated.

The data fidelity term quantifies how well estimated model parameters explain the measured data.
Our quadratic choice representing the negative log-likelihood under a Gaussian noise model is
\begin{align}
    \Ldata&(\mathbf{d},\mathbf{T},\bfemiss,\mathbf{\Omega}) 
    = \sum_{i = 1}^M\sum_{j = 1}^N\sum_{k = 1}^K (\yhat_{i,j,k}(\mathbf{d},\mathbf{T},\bfemiss, \mathbf{\Omega}) - y_{i,j,k})^2, 
\end{align}
where $\yhat_{i,j,k}$ and $y_{i,j,k}$ represent the model fit and the measured data, respectively, at spatial indices $i,j$ and spectral index $k$.
The model fit $\yhat_{i,j,k}$ is computed using the measurement model in~\eqref{eq:radiance_model_alt} and the reflection model in~\eqref{eq:discretized_reflected_light}.
Here, $M$ and $N$ denote the total number of vertical and horizontal spatial locations, respectively, and $K$ represents the total number of spectral measurements per spatial location.

Note that for $K$ spectral measurements, there are $K+Q+2$ unknowns for each pixel: 
$K$ from the emissivity profile,
$Q$ from projected solid angles,
and $2$ from object temperature and distance.
Minimization of the data fidelity term
is thus ill-posed, and regularization is needed to achieve a unique solution.
The emissivity profiles of solid objects are smooth in comparison to absorption lines\cite{Manolakis2019, borel2008error, borel2003artemiss}.  
To achieve physically plausible solutions, we use a regularizer that promotes smoothness in the emissivity estimates,
\begin{align}
    \Lreg(\bfemiss)
        &= \| \mathbf{D} \bfemiss \|_2^2 
        = \sum_{i = 1}^M\sum_{j = 1}^N\sum_{k = 1}^{K-1} (\emiss_{i,j,k+1} - \emiss_{i,j,k})^2,
\end{align}
where $\mathbf{D}$ is the operator that computes finite differences along the spectral dimension.

We additionally use total variation (TV) regularization for the distance parameter.
Unlike the regularization term for emissivity, the TV regularization is optional and is used for denoising purposes.
Specifically, we employ anisotropic TV regularization,
\begin{equation}
\text{TV}(\mathbf{d}) = \sum_{i = 1}^{M-1} \sum_{j = 1}^{N-1} \left( |d_{i+1,j} - d_{i,j}| + |d_{i,j+1} - d_{i,j}| \right).
\end{equation}

Combining the data fidelity and regularization terms, we solve the following optimization problem:
\begin{align}
  (\mathbf{d}^*&, \mathbf{T}^*, \bfemiss^*, \bm{\Omega}^*) \nonumber \\
  & = \argmin_{\mathbf{d},\mathbf{T},\bfemiss, \bm{\Omega}}
      \big( \Ldata(\mathbf{d},\mathbf{T},\bfemiss, \bm{\Omega}) 
      + \rho_{\emiss} \Lreg(\bfemiss) 
      + \rho_{d}\text{TV}(\mathbf{d}) \big)
\end{align}
subject to
\begin{subequations}
\begin{align}
          0 \leq d_{i,j} \leq \dmax, \quad &\forall i,j, \\
     0 \leq \emiss_{i,j,k} \leq 1, \quad &\forall i,j,k, \\
            \Omega_{i,j,q} \geq 0, \quad &\forall i,j,q, \\
   \sum_q \Omega_{i,j,q} \leq \pi, \quad &\forall i,j, 
\end{align}
\end{subequations}
where $\rho_{\emiss}$ is the regularization parameter for emissivity and $\rho_{d}$ is the regularization parameter for distance.
The constraints ensure the physical plausibility of the parameters;
the $\dmax$ upper bound on $d$ is optional but helps accelerate the optimization and should be set high enough to capture the full range of scene depths.
The optimization problem is solved using gradient descent while projecting the parameters on the constrained space through the iterations.

\section{Experimental Results}
\begin{figure*}
    \centering
    \begin{subfigure}[t]{0.2\textwidth}
        \centering
        \includegraphics[width=\linewidth]{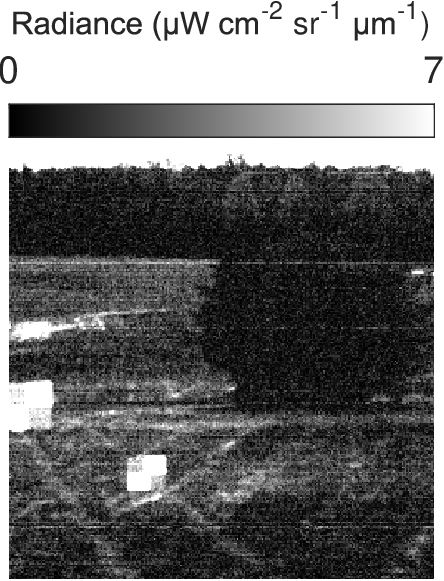}
        \caption{Difference around ozone \\ absorption}
        \label{fig:Experimental_bispectral_ozone}
    \end{subfigure}
    \hfill
    \begin{subfigure}[t]{0.2\textwidth}
        \centering
        \includegraphics[width=\linewidth]{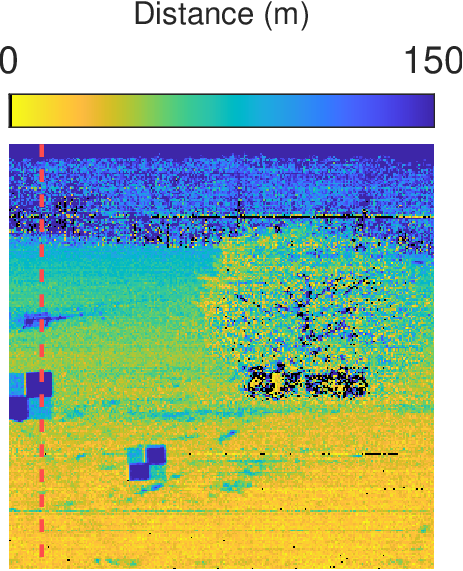}
        \caption{Bispectral (baseline)}
        \label{fig:Experimental_bispectral_baseline}
    \end{subfigure}
    \hfill
    \begin{subfigure}[t]{0.2\textwidth}
        \centering
        \includegraphics[width=\linewidth]{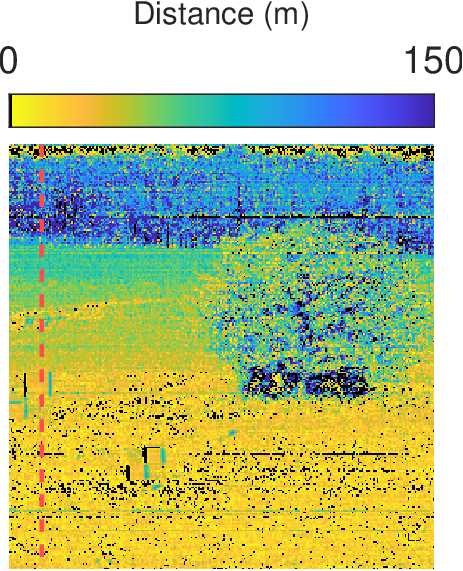}
        \caption{Quadspectral}
        \label{fig:Experimental_bispectral_ozone_correction}
    \end{subfigure}
    \hfill
    \begin{subfigure}[t]{0.2\textwidth}
        \centering
        \includegraphics[width=\linewidth]{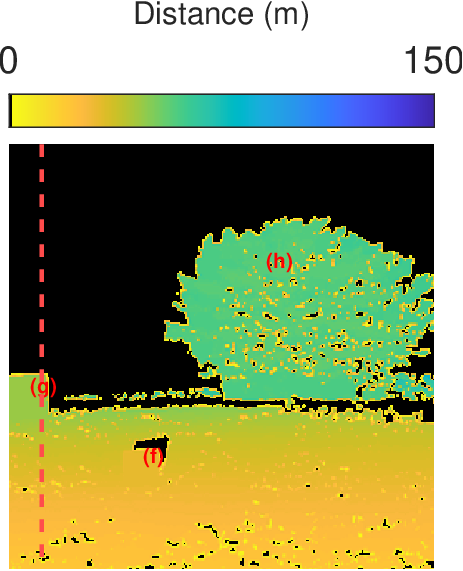}
        \caption{Lidar (ground truth)}
        \label{fig:Experimental_bispectral_lidar}
    \end{subfigure}
    \hfill\\
    \begin{subfigure}[t]{\textwidth}
        \centering
        \includegraphics[width=\linewidth]{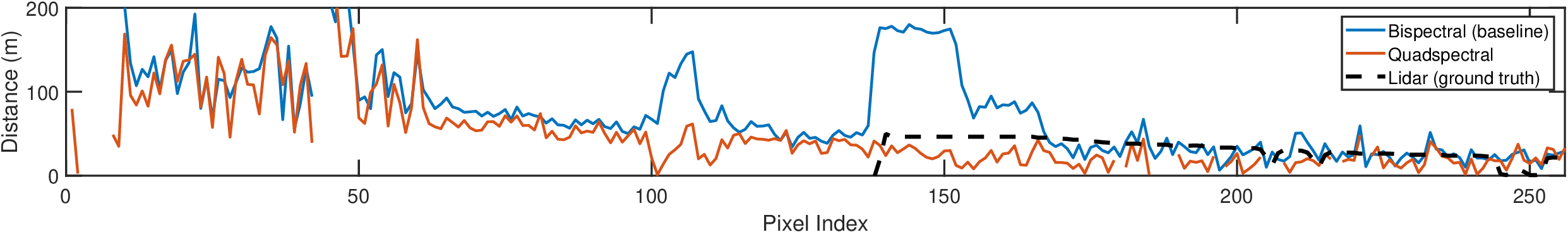}
        \caption{Vertical range profiles (highlighted with red vertical lines) of baseline bispectral method and ozone corrected bispectral method}
        \label{fig:Experimental_bispectral_comparison}
    \end{subfigure}
    \hfill\\
    \begin{subfigure}[t]{.3\textwidth}
        \centering
        \includegraphics[width=\linewidth]{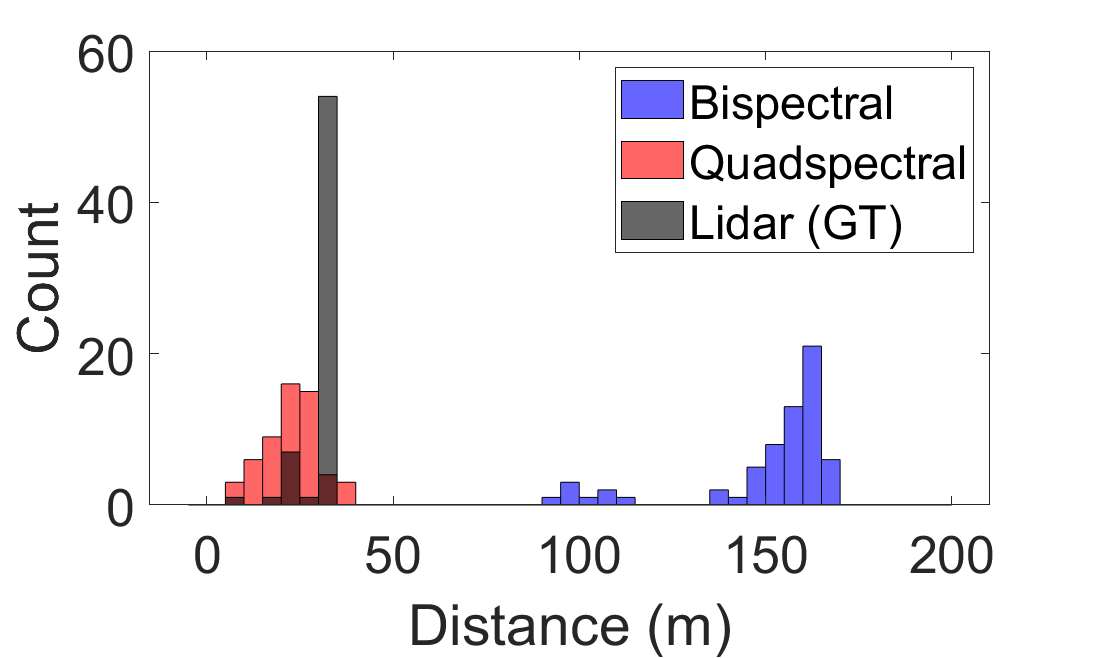}
        \caption{Histogram (front checkerboard)}
        \label{fig:bispectral_histogram_checkerboard_1}
    \end{subfigure}
    \hfill
    \begin{subfigure}[t]{.3\textwidth}
        \centering
        \includegraphics[width=\linewidth]{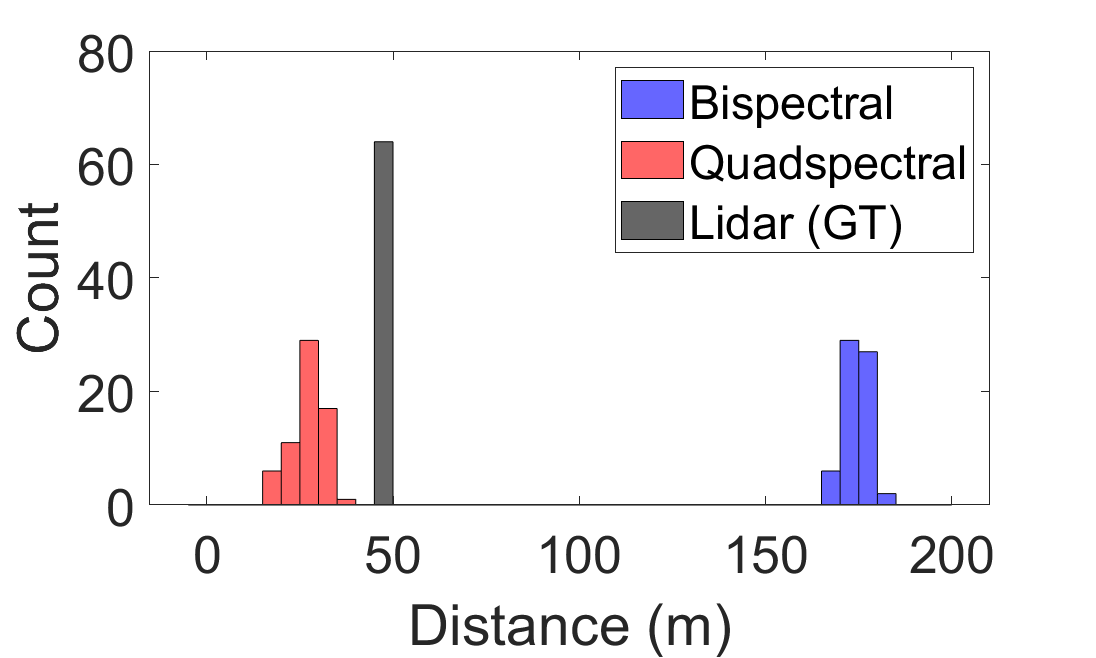}
        \caption{Histogram (rear checkerboard)}
        \label{fig:bispectral_histogram_checkerboard_2}
    \end{subfigure}
    \hfill
    \begin{subfigure}[t]{.3\textwidth}
        \centering
        \includegraphics[width=\linewidth]{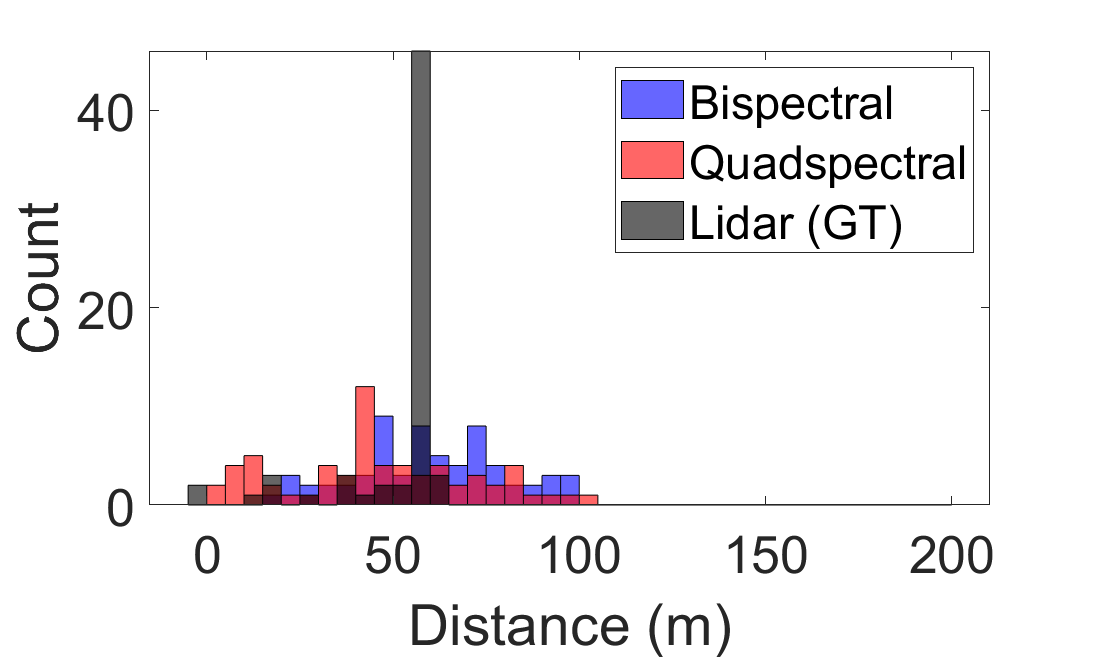}
        \caption{Histogram (tree)}
        \label{fig:bispectral_histogram_tree}
    \end{subfigure}
    \caption{Experimental results for bispectral ranging:
    \figref{fig:Experimental_bispectral_ozone}~difference of measurements around ozone absorption band; \figref{fig:Experimental_bispectral_baseline}~baseline bispectral method; \figref{fig:Experimental_bispectral_ozone_correction}~quadspectral method; \figref{fig:Experimental_bispectral_lidar}~lidar data approximately registered to hyperspectral LWIR sensor for ground truth;
    and
    \figref{fig:Experimental_bispectral_comparison} comparison of range profiles over the vertical line highlighted in red, with pixels indexed from top to bottom.
    Histograms of range estimates over $8 \times 8$ patches are shown in \figref{fig:bispectral_histogram_checkerboard_1} for the front checkerboard target; \figref{fig:bispectral_histogram_checkerboard_2} for the rear checkerboard target; and \figref{fig:bispectral_histogram_tree} for the tree; the corresponding locations are indicated in the lidar map.
    The black pixels in the depth map represent the pixels with invalid estimations or no lidar data. 
    The baseline bispectral method is vulnerable to reflected downwelling radiation resulting in overestimations for reflective objects.
    Including two more absorption lines around the ozone band and following the quadspectral method
    gives much more reliable results.}
    \label{fig:Experimental_bispectral_results}
\end{figure*}

\subsection{Data Specifications}
The experimental data provided by the U.S. Army Night Vision and Electronic Sensors Directorate and the Johns Hopkins University Applied Physics Laboratory were acquired using a pushbroom LWIR hyperspectral imager with a cooled HgCdTe sensor.
The same data as~\cite{10877411} is used and it is publicly available in~\cite{Yellin_2024_WACV}.
The spectrometer has a vertical field-of-view (VFOV) of 11.6 degrees and a horizontal FOV (HFOV) of 57 degrees.
The focal length is 50 mm and the f-number is f/0.9\@.
The typical noise of the sensor is around 1 microflick, which---at 10\,\si{\micro \meter} wavelength---corresponds to about a 1000:1 signal-to-noise ratio~\cite{bao2023heat}.
For each of the $1280 \times 260$ pixels in the scene, the sensor acquires data for 256 spectral bands between 8.0\,\si{\micro\meter} and 13.2\,\si{\micro\meter}\@. The spectral resolution is $\approx 40\,\si{\nano\meter}$.
Shifting raw data by $-120\,\si{nm}$
was found to be a necessary calibration for consistency with
known atmospheric absorption features.

During the data collection, an on-site weather station recorded humidity, air temperature and pressure levels.
These three parameters were used to calculate the attenuation function with a high-resolution spectral modeling software (Spectralcalc)~\cite{SpectralCalc}.
We assume these parameters are constant over the scene and use a single attenuation function and air temperature for all pixels.
Lidar measurements were also acquired using a high-resolution system that collects 1240 points over 360$^{\circ}$, with a point spacing of $\approx 6$\,\si{\milli\meter} at 10\,\si{\meter}, a VFOV of 150$^{\circ}$, and an HFOV of $\approx 90^{\circ}$;
the system has ranging uncertainty of $\pm$1\,\si{\milli\meter} and a maximum range of 350\,\si{\meter}.

The experimental data
that we study extensively
were collected near sunset,
19:41 local time on April 13, 2021,
and labeled as Path 5, Step 1
in dataset IHTest\_202104~\cite{Yellin_2024_WACV}.
The scene mostly consists of rolling grassy terrain.
There are sky and forest pixels at the top of the VFOV\@.
There is a tree in the right foreground of the scene, and
two reflective panels with checkerboard patterns on the left.

\subsection{Quadspectral Results}

The ranging results for the quadspectral method are presented in \cref{fig:Experimental_bispectral_results}.
\Cref{fig:Experimental_bispectral_ozone} illustrates the absolute difference across the ozone absorption band in the scene. 
The ozone feature in the measurements highlights the contribution of downwelling radiance, as discussed in Section~\ref{sec:ozone_absorption}.
As a baseline, \cref{fig:Experimental_bispectral_baseline} shows the result of bispectral method~\eqref{eq:bispectral_air} which accounts for air emission but does not account for downwelling radiance, using selected wavelengths of $\lambda_1 = 8.42\,\si{\micro\meter}$, $\lambda_2 = 8.46\,\si{\micro\meter}$.
Neglecting the downwelling radiance affects various parts of the scene, depending on the emissivity and orientation of each object. 
The checkerboard targets, being highly reflective, are significantly overestimated.
In contrast, the grassy areas, which generally exhibit high emissivity (and thus low reflectivity), are minimally affected, except for some dry patches and wheel-tracked regions.
Ozone features are used to correct the overestimations using the quadspectral estimation, as detailed in Section~\ref{sec:quadspectral_estimation}.
The quadspectral results are presented in \cref{fig:Experimental_bispectral_ozone_correction}, using selected wavelengths of $\lambda_1 = 8.42\,\si{\micro\meter}$, $\lambda_2 = 8.46\,\si{\micro\meter}$, $\lambda_3 = 9.49\,\si{\micro\meter}$, and $\lambda_4 = 9.57\,\si{\micro\meter}$.
Improvement of the quadspectral method over the bispectral method is evident.
\Cref{fig:Experimental_bispectral_comparison} compares the baseline bispectral method~\eqref{eq:bispectral_air} with the quadspectral estimation along a vertical line from the top to the bottom of the image, where artifacts are most prominent.
Around pixel index 150, the checkerboard target, initially overestimated by the baseline method, is significantly improved with ozone correction applied.
Overall, a negative bias is observed relative to the lidar measurements, likely resulting from assumptions about atmospheric properties, such as air temperature and water vapor content, or from discrepancies between the modeled and actual downwelling radiance, which is based on a standard atmospheric model~\cite{coesa1976standard}.

The lidar map is only approximately registered to the hyperspectral camera and is therefore not suitable for pixel-wise performance evaluation across the entire scene.
Instead, we compare depth estimates over non-overlapping $8 \times 8$ patches at representative locations, highlighted in \cref{fig:Experimental_bispectral_lidar}.
These patches are chosen to be spatially homogeneous in ground-truth depth while large enough (64 pixels) to allow statistical evaluation.
Histograms of the estimated depth values for different methods are shown in \cref{fig:bispectral_histogram_checkerboard_1} for the front checkerboard,
\cref{fig:bispectral_histogram_checkerboard_2} for the rear checkerboard, and \cref{fig:bispectral_histogram_tree} for the tree.
The corresponding means and standard deviations are reported in \cref{tab:Experimental_bispectral_table}.
The median of the lidar measurements is given as the ground truth proxy.
For the reflective checkerboards, the quadspectral method yields significantly more accurate and consistent results than the baseline bispectral method.
In contrast, only minimal improvement is observed in the less reflective tree region, due to the low contribution of downwelling radiance.

\begin{table}
\caption{Means and standard deviations of experimental quadspectral results on $8 \times 8$ patches; histograms shown in Fig.~\ref{fig:Experimental_bispectral_results}.}
\label{tab:Experimental_bispectral_table}
\centering
\scalebox{0.77}{
\begin{tabular}{@{}rr@{\,}c@{\,}rr@{\,}c@{\,}rr@{\,}c@{\,}r@{}}
\toprule
Method &
\multicolumn{3}{r}{Front Checkerboard} &
\multicolumn{3}{r}{Rear Checkerboard} &
\multicolumn{3}{r@{}}{Tree} \\
\midrule
Bispectral    & 150.73\,\si{\meter} & $\pm$ & 19.89\,\si{\meter}
              & 174.26\,\si{\meter} & $\pm$ &  3.05\,\si{\meter}
              &  59.83\,\si{\meter} & $\pm$ & 20.14\,\si{\meter} \\
Quadspectral  &  22.84\,\si{\meter} & $\pm$ &  7.20\,\si{\meter}
              &  26.77\,\si{\meter} & $\pm$ &  4.63\,\si{\meter}
              &  46.47\,\si{\meter} & $\pm$ & 25.33\,\si{\meter} \\
Ground truth proxy    &  31.01\,\si{\meter} &       & 
              &  46.51\,\si{\meter} &       & 
              &  57.84\,\si{\meter} &       &  \\
\bottomrule
\end{tabular}
}
\end{table}

\subsection{Hyperspectral Results}

\begin{figure*}
    \centering
    \begin{subfigure}[t]{0.2\textwidth}
        \centering
        \includegraphics[width=\linewidth]{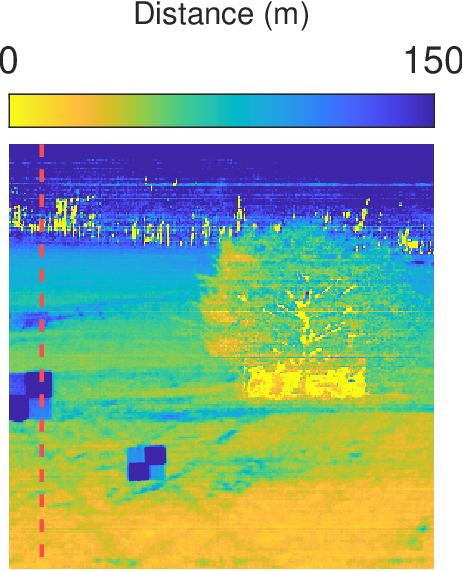}
        \caption{Hyperspectral (neglecting \\ downwelling)}
        \label{fig:Experimental_hyperspectral_results_range_nd}
    \end{subfigure}
    \hfill
    \begin{subfigure}[t]{0.2\textwidth}
        \centering
        \includegraphics[width=\linewidth]{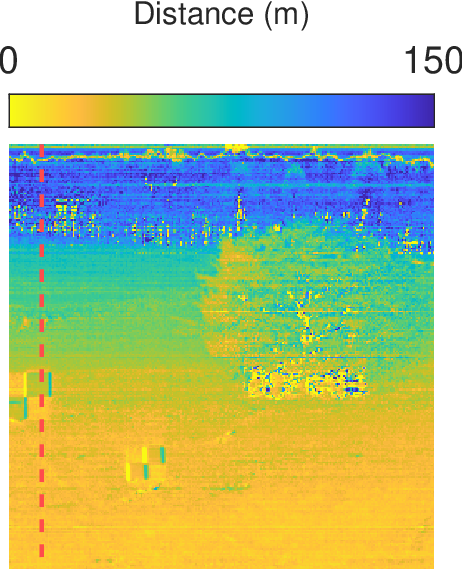}
        \caption{Hyperspectral (including \\ downwelling)}
        \label{fig:Experimental_hyperspectral_results_range_wd}
    \end{subfigure}
    \hfill
    \begin{subfigure}[t]{0.2\textwidth}
        \centering
        \includegraphics[width=\linewidth]{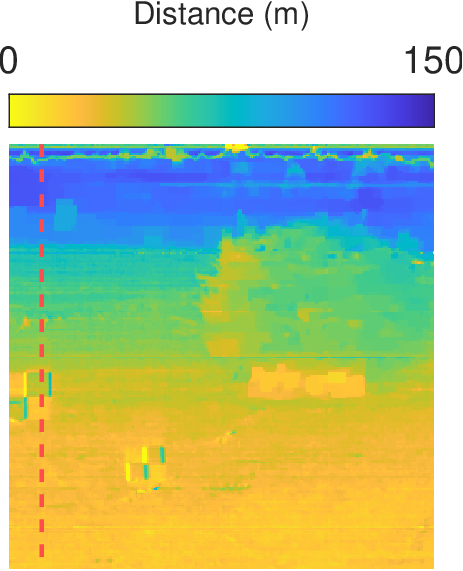}
        \caption{Hyperspectral-TV (including \\ downwelling)}
        \label{fig:Experimental_hyperspectral_results_range_wd_TV}
    \end{subfigure}
    \hfill
    \begin{subfigure}[t]{0.2\textwidth}
        \centering
        \includegraphics[width=\linewidth]{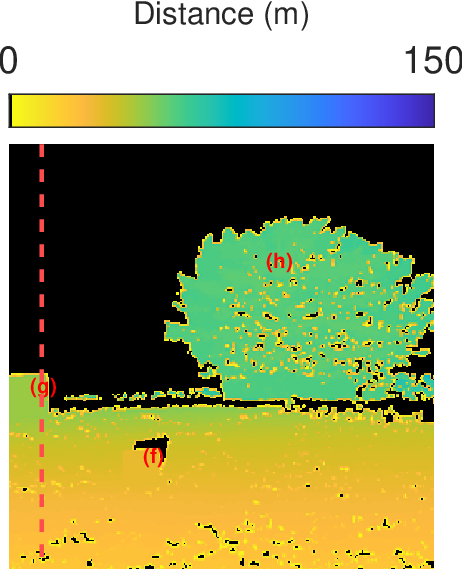}
        \caption{Lidar (ground truth)}
        \label{fig:Experimental_hyperspectral_results_lidar}
    \end{subfigure}
    \hfill \\
    \begin{subfigure}[t]{\textwidth}
        \centering
        \includegraphics[width=\linewidth]{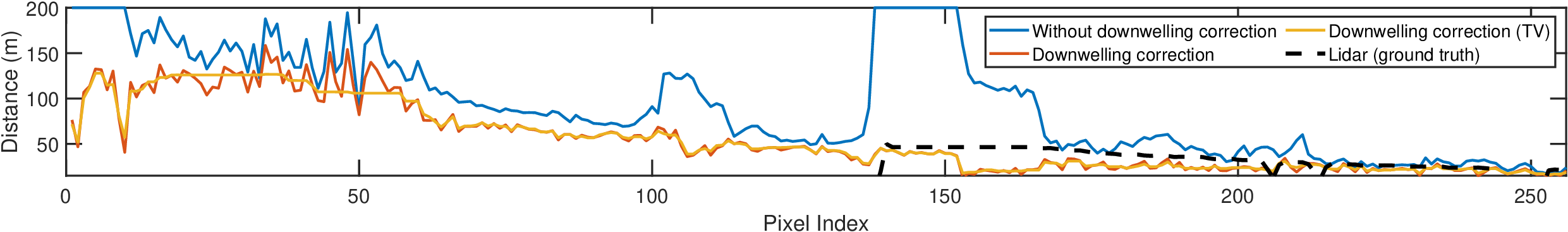}
        \caption{Vertical range profiles (highlighted with red vertical lines) of hyperspectral methods neglecting and including downwelling radiance}
        \label{fig:Experimental_hyperspectral_results_comparison}
    \end{subfigure}
    \hfill \\
    \begin{subfigure}[t]{.3\textwidth}
        \centering
        \includegraphics[width=\linewidth]{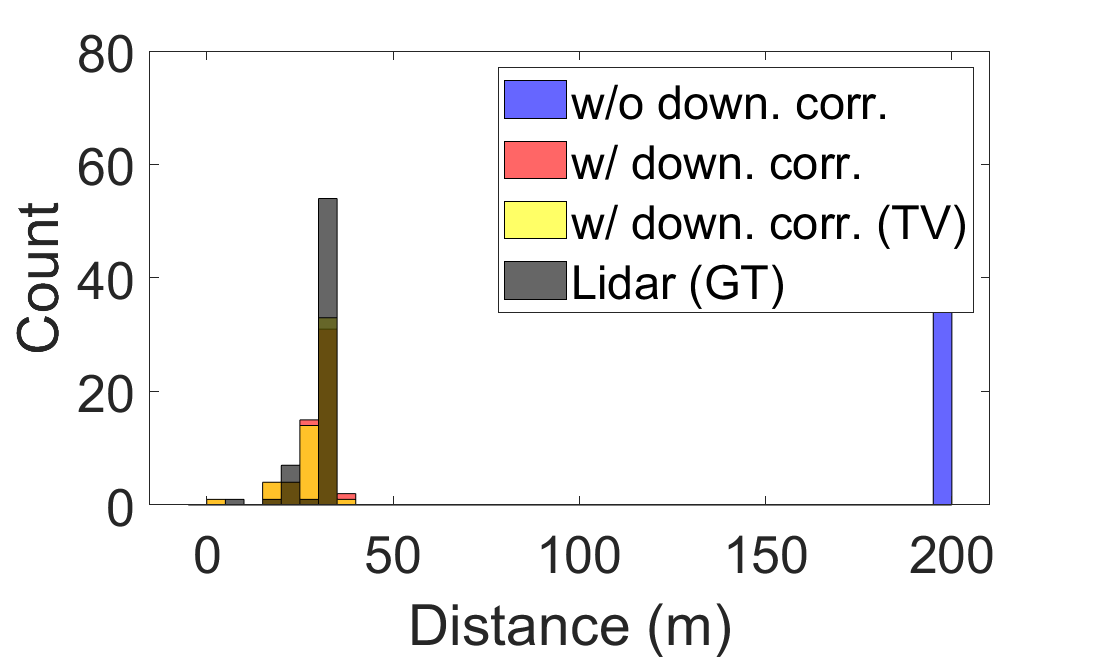}
        \caption{Histogram (front checkerboard)}
        \label{fig:hyperspectral_histogram_checkerboard_1}
    \end{subfigure}
    \hfill
    \begin{subfigure}[t]{.3\textwidth}
        \centering
        \includegraphics[width=\linewidth]{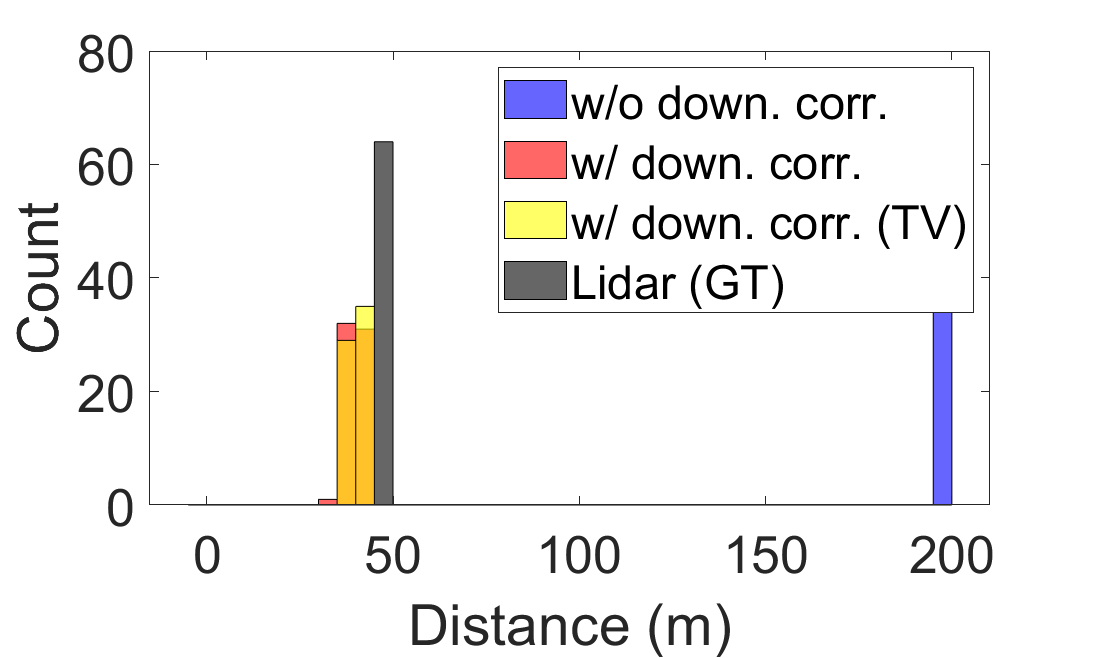}
        \caption{Histogram (rear checkerboard)}
        \label{fig:hyperspectral_histogram_checkerboard_2}
    \end{subfigure}
    \hfill
    \begin{subfigure}[t]{.3\textwidth}
        \centering
        \includegraphics[width=\linewidth]{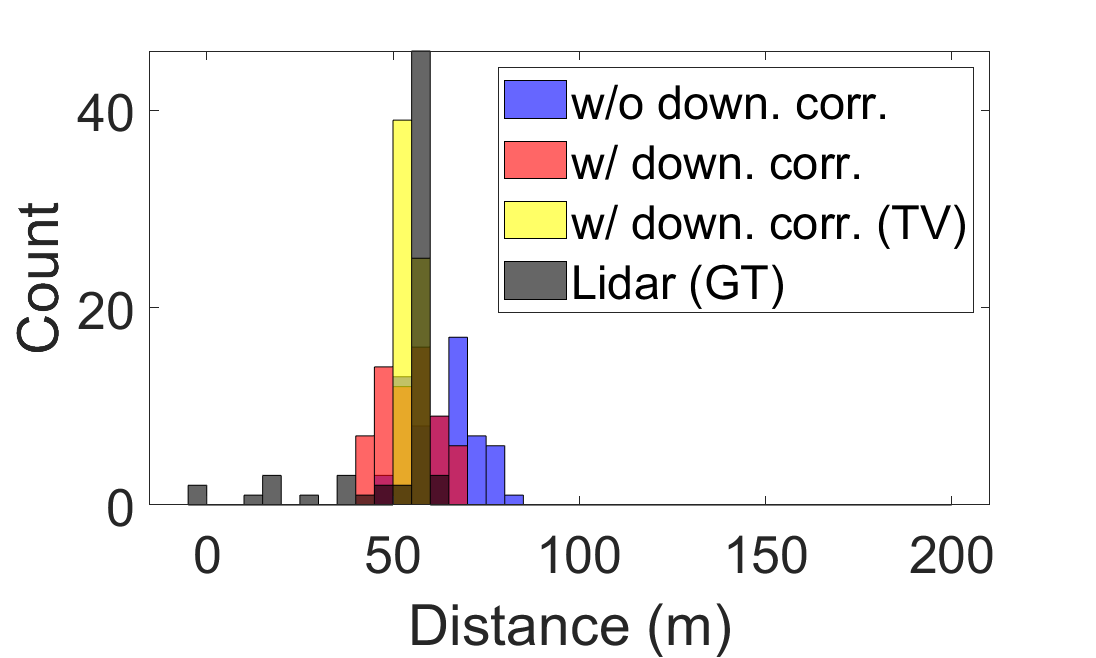}
        \caption{Histogram (tree)}
        \label{fig:hyperspectral_histogram_tree}
    \end{subfigure}
    \caption{Experimental results for hyperspectral ranging: \figref{fig:Experimental_hyperspectral_results_range_nd} hyperspectral method neglecting reflected downwelling radiance;
    \figref{fig:Experimental_hyperspectral_results_range_wd} hyperspectral method accounting for reflected downwelling radiance, without TV regularization;
    \figref{fig:Experimental_hyperspectral_results_range_wd_TV} hyperspectral method accounting for reflected downwelling radiance, with TV regularization;
    \figref{fig:Experimental_hyperspectral_results_lidar} lidar data registered to hyperspectral LWIR sensor for ground truth;
    and
    \figref{fig:Experimental_hyperspectral_results_comparison} comparison vertical range profiles highlighted with red dashed line, with pixels indexed from top to bottom.
    Histograms of range estimates over $8 \times 8$ patches are shown in \figref{fig:hyperspectral_histogram_checkerboard_1} for the front checkerboard target; \figref{fig:hyperspectral_histogram_checkerboard_2} for the rear checkerboard target; and \figref{fig:hyperspectral_histogram_tree} for the tree;
    the corresponding locations are indicated in the lidar map.
    The black pixels in the depth map represent the pixels with invalid estimations or no lidar data.
    Including downwelling radiation in the model significantly removes the overestimations around pixel indices 100 and 150.}
    \label{fig:Experimental_hyperspectral_results}
\end{figure*}

Ranging results for the hyperspectral method are provided in \cref{fig:Experimental_hyperspectral_results}.
The results of hyperspectral ranging without accounting for reflected downwelling radiance are presented in \cref{fig:Experimental_hyperspectral_results_range_nd} while \cref{fig:Experimental_hyperspectral_results_range_wd} illustrates the depth map accounting for downwelling radiance incident at different zenith angles.
The maximum distance $\dmax$ is set to 200\,\si{\meter}.
Similar to the quadspectral method, including downwelling radiance reduces the overestimates across the scene.
Notably, the overestimated areas around pixel indices 100 and 150 are improved when downwelling radiance is accounted for.
While this correction leads to substantially more reliable estimates overall, some residual errors remain, such as the discontinuity near pixel 150, indicating that the decomposition is not perfect.
Again, a consistent negative bias relative to lidar measurements is also observed, which is likely due to assumptions regarding atmospheric parameters, including air temperature, water vapor content, or the use of a standard atmospheric model for simulating downwelling radiance~\cite{coesa1976standard}.
The TV regularized version of the hyperspectral estimation is shown in \cref{fig:Experimental_hyperspectral_results_range_wd_TV}.
The TV regularization affects the reconstruction based on the noise level.
For example the background forest that is initially noisier due to low temperature contrast is more affected by regularization compared to other parts of the scene.  
\Cref{fig:Experimental_hyperspectral_results_comparison} displays the range profiles along the red vertical line from the top to the bottom of the scene for better comparison. 

Numerical comparisons are performed over identical $8 \times 8$ patches. Depth estimate histograms are shown in \cref{fig:hyperspectral_histogram_checkerboard_1} for the front checkerboard, \cref{fig:hyperspectral_histogram_checkerboard_2} for the rear checkerboard, and \cref{fig:hyperspectral_histogram_tree} for the tree region. 
Corresponding means and standard deviations are summarized in \cref{tab:Experimental_hyperspectral_table}.
For cases where the hyperspectral estimates reached the upper bound of 200\,\si{\meter}, the standard deviation is reported as zero.
Note the improved accuracy of estimates when downwelling correction in included.
Furthermore, comparing to \cref{tab:Experimental_bispectral_table}
shows that the hyperspectral methods have better accuracy and consistency than the quadspectral method;
as more wavelengths are incorporated, the influence of noise on the depth estimates diminishes.

\begin{figure}
    \centering
    \begin{subfigure}[t]{0.45\linewidth}
        \centering
        \includegraphics[width=\linewidth]{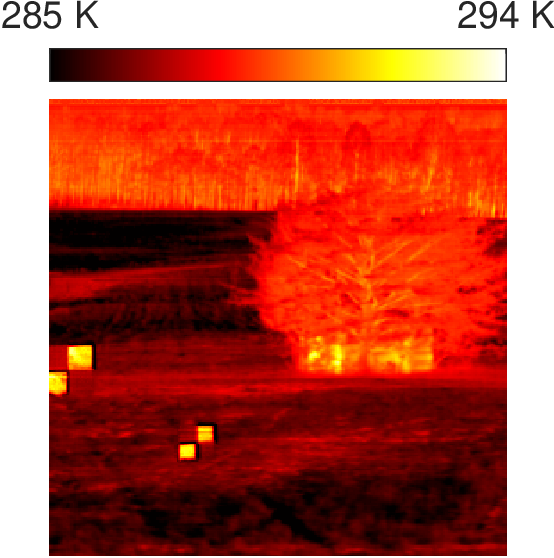}
        \caption{Temperature (neglecting \\ downwelling)}
        \label{fig:Experimental_hyperspectral_temperature_nd}
    \end{subfigure}
    \hfill
    \begin{subfigure}[t]{0.45\linewidth}
        \centering
        \includegraphics[width=\linewidth]{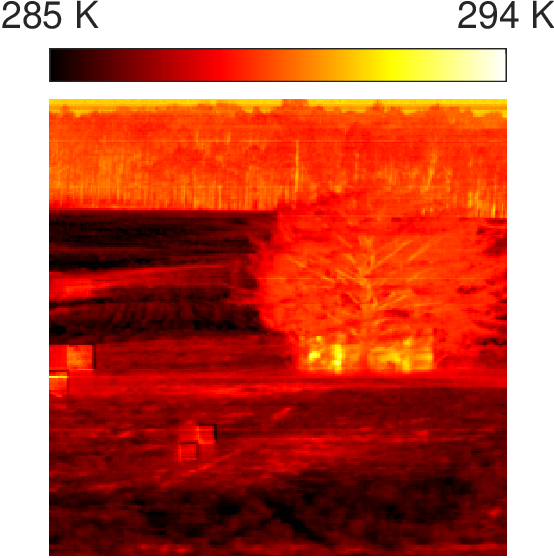}
        \caption{Temperature (including \\ downwelling)}
        \label{fig:Experimental_hyperspectral_temperature_wd}
    \end{subfigure}
    \hfill \\
    \begin{subfigure}[t]{0.45\linewidth}
        \centering
        \includegraphics[width=\linewidth]{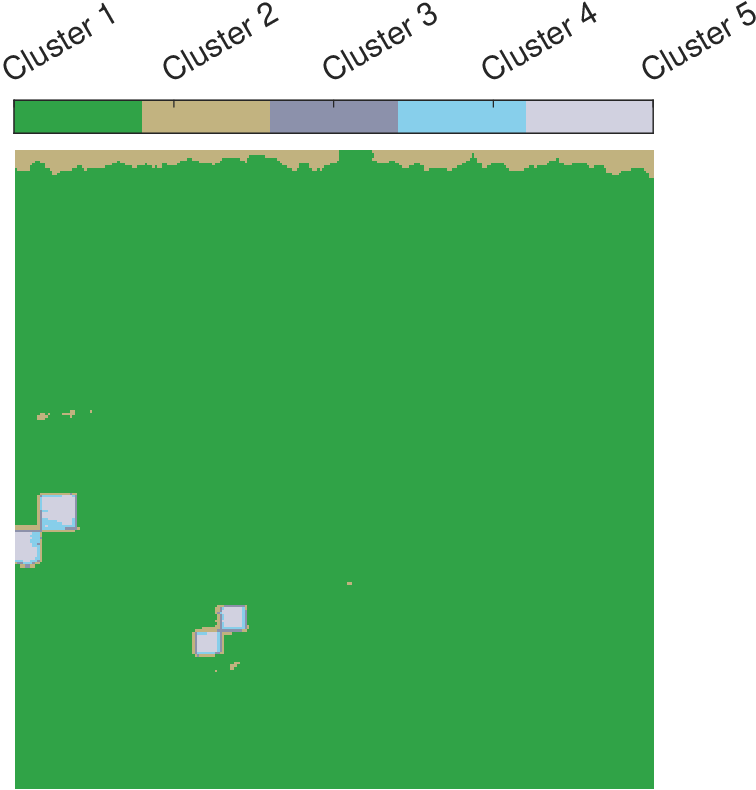}
        \caption{Emissivity cluster map \\ (neglecting downwelling)}
        \label{fig:Experimental_hyperspectral_emissivity_map_nd}
    \end{subfigure}
    \hfill
    \begin{subfigure}[t]{0.45\linewidth}
        \centering
        \includegraphics[width=\linewidth]{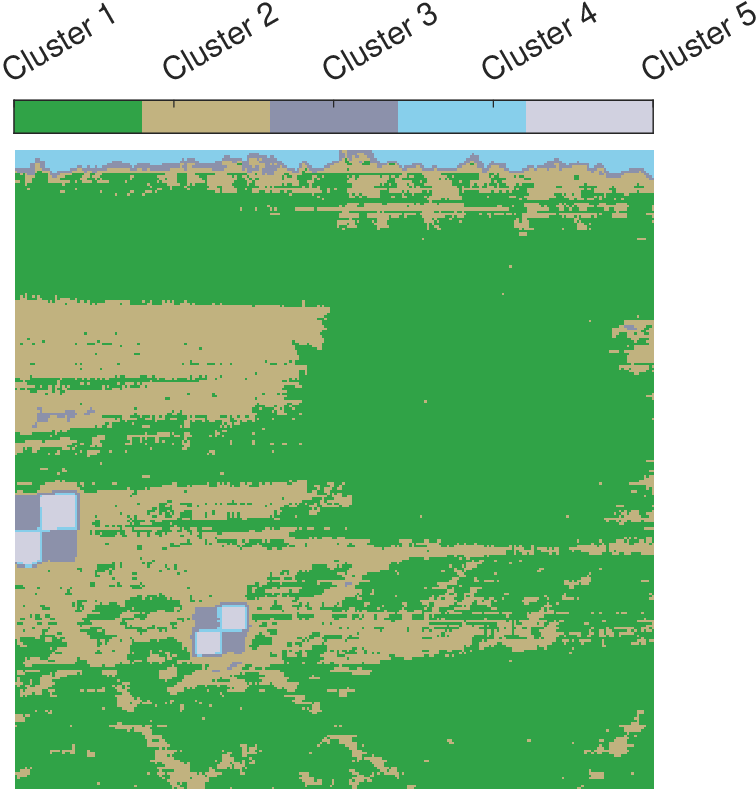}
        \caption{Emissivity cluster map \\ (including downwelling)}
        \label{fig:Experimental_hyperspectral_emissivity_map_wd}
    \end{subfigure}
    \hfill \\
    \begin{subfigure}[t]{0.45\linewidth}
        \centering
        \includegraphics[width=\linewidth]{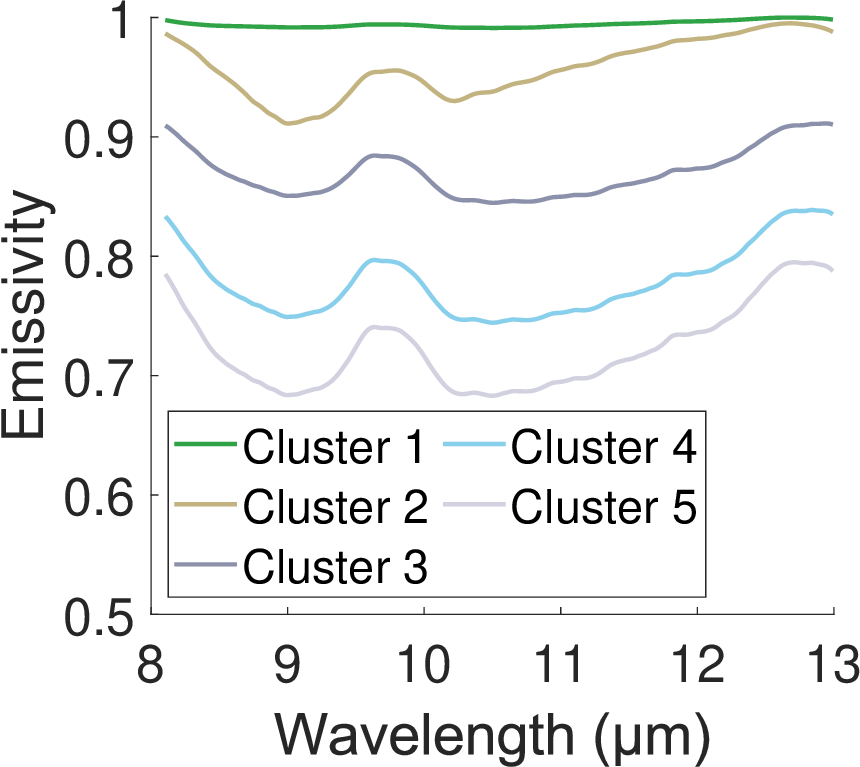}
        \caption{Emissivity cluster spectra \\ (neglecting downwelling)}
        \label{fig:Experimental_hyperspectral_emissivity_spectra_nd}
    \end{subfigure}
    \hfill
    \begin{subfigure}[t]{0.45\linewidth}
        \centering
        \includegraphics[width=\linewidth]{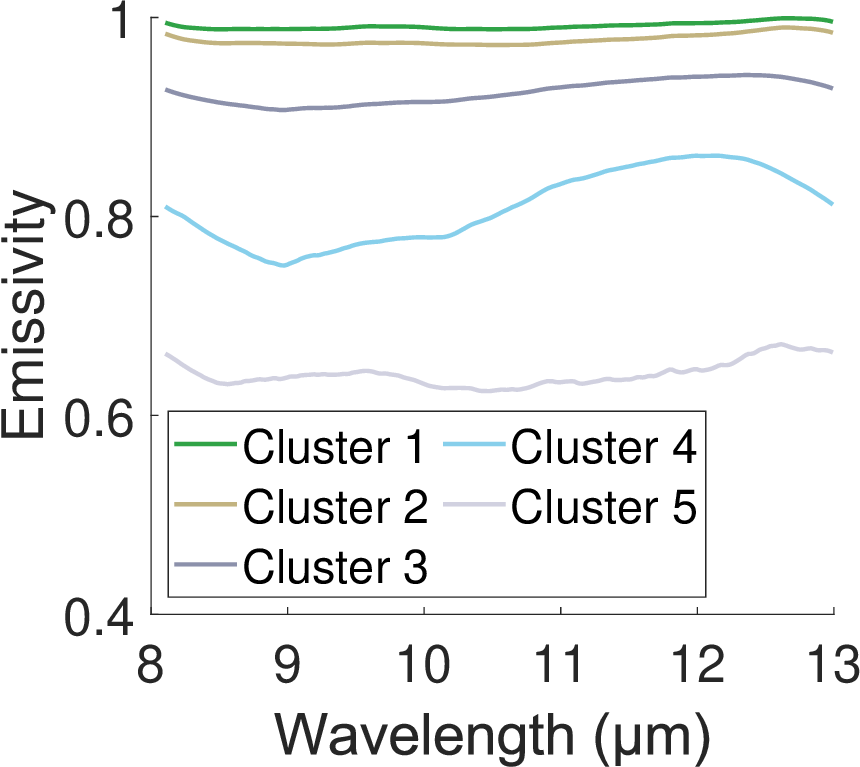}
        \caption{Emissivity cluster spectra \\ (including downwelling)}
        \label{fig:Experimental_hyperspectral_emissivity_spectra_wd}
    \end{subfigure}
    \caption{Temperature and emissivity results for hyperspectral method:
    \figref{fig:Experimental_hyperspectral_temperature_nd} and \figref{fig:Experimental_hyperspectral_temperature_wd} shows the temperature estimates for neglecting and including downwelling radiance, respectively;
    \figref{fig:Experimental_hyperspectral_emissivity_map_nd} and \figref{fig:Experimental_hyperspectral_emissivity_map_wd} shows emissivity clusters neglecting and including downwelling radiance, respectively; and
    \figref{fig:Experimental_hyperspectral_emissivity_spectra_nd} and \figref{fig:Experimental_hyperspectral_emissivity_spectra_wd} shows the mean spectra for the clusters.
    K-means with 5 groups is used to generate the emissivity clusters.} 
    \label{fig:Experimental_hyperspectral_results_temperature_emissivity}
\end{figure}

\begin{table}
\caption{Means and standard deviations of experimental hyperspectral results on $8 \times 8$ patches; histograms shown in \cref{fig:Experimental_hyperspectral_results}.}
\label{tab:Experimental_hyperspectral_table}
\centering
\scalebox{0.77}{
\begin{tabular}{@{}rr@{\,}c@{\,}rr@{\,}c@{\,}rr@{\,}c@{\,}r@{}}
\toprule
Method &
\multicolumn{3}{r}{Front Checkerboard} &
\multicolumn{3}{r}{Rear Checkerboard} &
\multicolumn{3}{r@{}}{Tree} \\
\midrule
without down.\ corr.\    & 200.00\,\si{\meter} & $\pm$ &  0.00\,\si{\meter}
                     & 200.00\,\si{\meter} & $\pm$ &  0.00\,\si{\meter}
                     &  63.00\,\si{\meter} & $\pm$ &  8.85\,\si{\meter} \\
with down.\ corr.\      &  28.86\,\si{\meter} & $\pm$ &  5.38\,\si{\meter}
                     &  39.83\,\si{\meter} & $\pm$ &  2.01\,\si{\meter}
                     &  54.27\,\si{\meter} & $\pm$ &  7.40\,\si{\meter} \\
with down.\ corr.\ (TV)  &  28.82\,\si{\meter} & $\pm$ &  5.31\,\si{\meter}
                     &  39.84\,\si{\meter} & $\pm$ &  1.78\,\si{\meter}
                     &  53.99\,\si{\meter} & $\pm$ &  2.01\,\si{\meter} \\
Ground truth proxy           &  31.01\,\si{\meter} &  &
                     &  46.51\,\si{\meter} &       &
                     &  57.84\,\si{\meter} &       & \\
\bottomrule
\end{tabular}
}
\end{table}

Though our emphasis is on ranging, we conclude with a discussion of the auxiliary estimates produced by the hyperspectral method.
The temperature and emissivity estimate results are shown in \cref{fig:Experimental_hyperspectral_results_temperature_emissivity}.
\Cref{fig:Experimental_hyperspectral_temperature_nd,fig:Experimental_hyperspectral_temperature_wd} show the temperature estimate neglecting and accounting for the downwelling radiance, respectively.
Not accounting for the downwelling radiance also causes errors in estimating the temperature especially for highly reflective objects.
This effect is most notable for the reflective parts of the checkerboard target.
Without any heat source, an object is expected to have a nearly uniform temperature due to heat conduction~\cite{incropera1996fundamentals}.
However when downwelling is not accounted for, a temperature difference can be seen in the checkerboards.
When the downwelling radiance is accounted for the temperature distribution over the checkerboards become uniform as it is expected.

Emissivity estimates are computed per pixel and wavelength, forming a spectral datacube. To visualize and analyze variations in emissivity profiles across the scene, we apply k-means clustering~\cite{macqueen1967some}, treating each pixel as a data point and its emissivity profile (across wavelengths) as the feature vector.
We specify the number of clusters as five, allowing us to group the raw emissivity estimates into five representative emissivity profiles.
\Cref{fig:Experimental_hyperspectral_emissivity_map_nd,fig:Experimental_hyperspectral_emissivity_map_wd} show the map of the emissivity clusters neglecting and accounting for downwelling radiance, respectively.
Accounting for downwelling radiance also improves classification of objects.
Dry grass and soil areas become distinguishable, accounting for downwelling radiance, as well as different parts of the checkerboard target.
\Cref{fig:Experimental_hyperspectral_emissivity_spectra_nd,fig:Experimental_hyperspectral_emissivity_spectra_wd} show the mean spectra of these five clusters when neglecting and accounting for downwelling radiance. 
The color codes are the same as the emissivity cluster maps for matching.
Not accounting for downwelling radiance shows significantly different emissivity profiles, especially at 9.5\,\si{\micro\meter} where ozone absorption is present.
The emissivity fit neglecting downwelling radiance shows artifacts at the ozone absorption area (smoothed by the regularizer) to represent the radiance variation in the measurements.
Considering downwelling radiance corrects artifacts in the emissivity estimates caused by ozone absorption.

\begin{figure}
    \centering
    \begin{subfigure}[t]{0.15\textwidth}
        \centering
        \includegraphics[width=\linewidth]{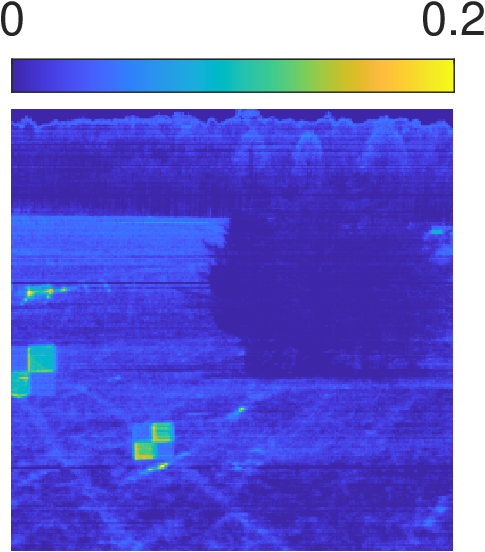}
        \caption{$0^{\circ}$}
        \label{fig:Experimental_hyperspectral_normalized_solid_angle_0deg}
    \end{subfigure}
    \hfill
    \begin{subfigure}[t]{0.15\textwidth}
        \centering
        \includegraphics[width=\linewidth]{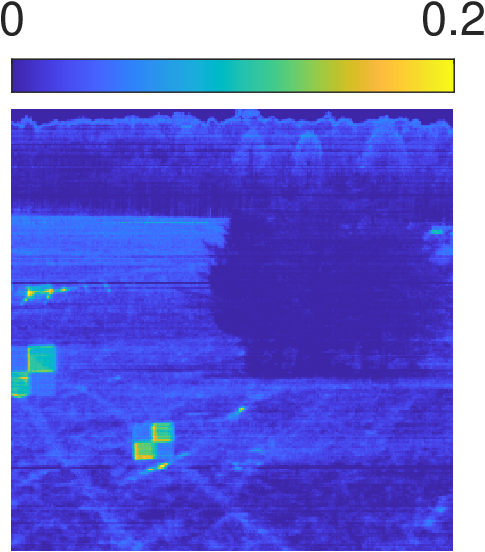}
        \caption{$30^{\circ}$}
        \label{fig:Experimental_hyperspectral_normalized_solid_angle_30deg}
    \end{subfigure}
    \hfill
    \begin{subfigure}[t]{0.15\textwidth}
        \centering
        \includegraphics[width=\linewidth]{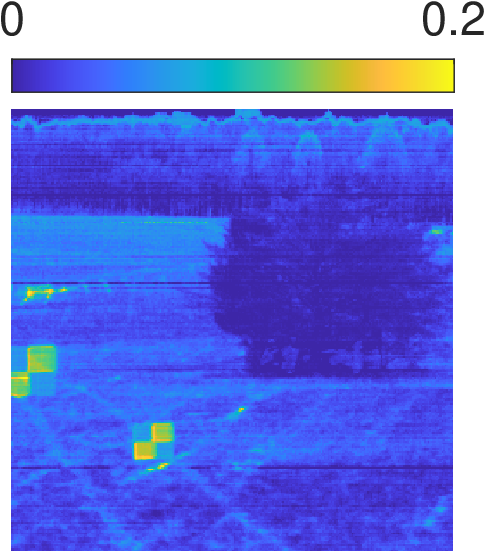}
        \caption{$60^{\circ}$}
        \label{fig:Experimental_hyperspectral_normalized_solid_angle_60deg}
    \end{subfigure}
    \hfill
    \begin{subfigure}[t]{0.15\textwidth}
        \centering
        \includegraphics[width=\linewidth]{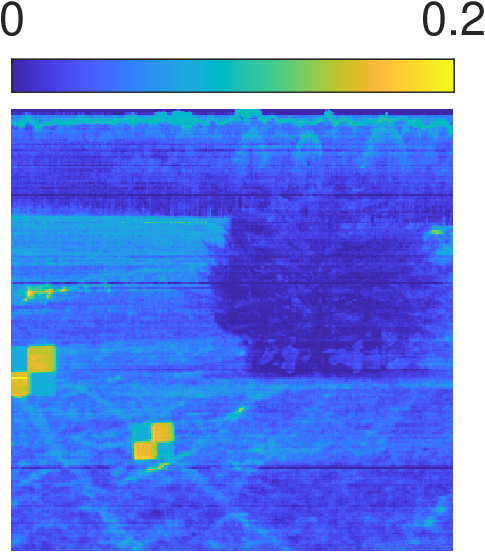}
        \caption{$70^{\circ}$}
        \label{fig:Experimental_hyperspectral_normalized_solid_angle_70deg}
    \end{subfigure}
    \hfill
    \begin{subfigure}[t]{0.15\textwidth}
        \centering
        \includegraphics[width=\linewidth]{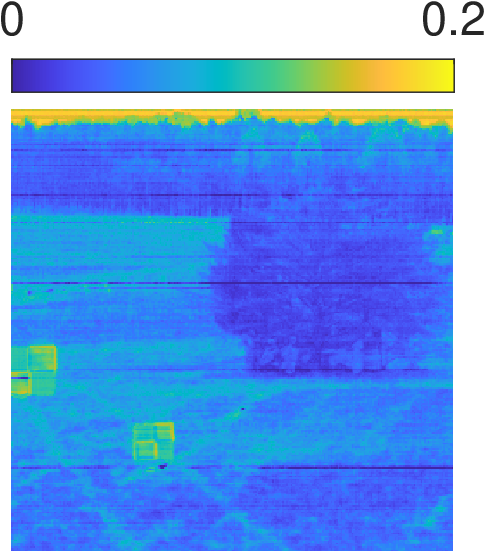}
        \caption{$80^{\circ}$}
        \label{fig:Experimental_hyperspectral_normalized_solid_angle_80deg}
    \end{subfigure}
    \hfill
    \begin{subfigure}[t]{0.15\textwidth}
        \centering
        \includegraphics[width=\linewidth]{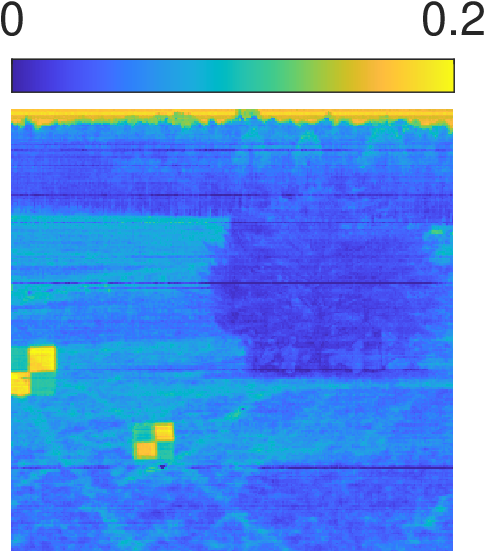}
        \caption{$82^{\circ}$}
        \label{fig:Experimental_hyperspectral_normalized_solid_angle_82deg}
    \end{subfigure}
    \hfill
    \begin{subfigure}[t]{0.15\textwidth}
        \centering
        \includegraphics[width=\linewidth]{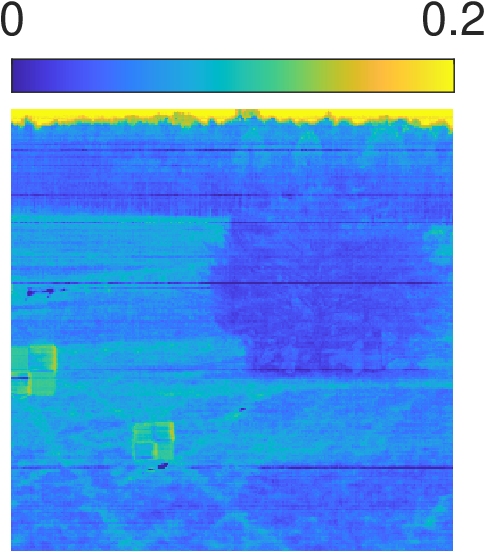}
        \caption{$84^{\circ}$}
        \label{fig:Experimental_hyperspectral_normalized_solid_angle_84deg}
    \end{subfigure}
    \hfill
    \begin{subfigure}[t]{0.15\textwidth}
        \centering
        \includegraphics[width=\linewidth]{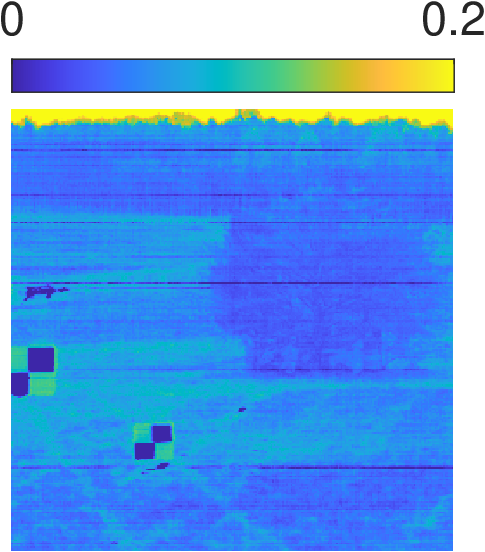}
        \caption{$86^{\circ}$}
        \label{fig:Experimental_hyperspectral_normalized_solid_angle_86deg}
    \end{subfigure}
    \hfill
    \begin{subfigure}[t]{0.15\textwidth}
        \centering
        \includegraphics[width=\linewidth]{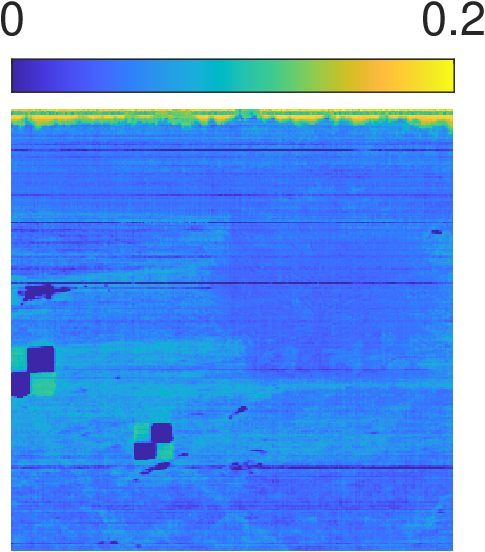}
        \caption{$88^{\circ}$}
        \label{fig:Experimental_hyperspectral_normalized_solid_angle_88deg}
    \end{subfigure}
    \hfill
    \begin{subfigure}[t]{0.15\textwidth}
        \centering
        \includegraphics[width=\linewidth]{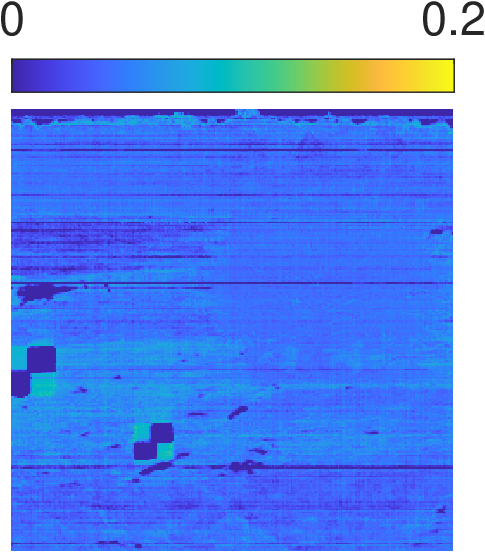}
        \caption{$89^{\circ}$}
        \label{fig:Experimental_hyperspectral_normalized_solid_angle_89deg}
    \end{subfigure}
    \caption{Normalized and projected solid angle estimates ($\Omega_i/\pi$) in the hyperspectral method.
    The values show the weight of downwelling radiances at various angles from $0^{\circ}$ to $90^{\circ}$ zenith angles.
    The combination of downwelling radiances with estimated weights forms the reflected radiance component in the model.}
\label{fig:Experimental_hyperspectral_result_solid_angles}
\end{figure}

\Cref{fig:Experimental_hyperspectral_result_solid_angles} shows estimated normalized and projected solid angles ($\Omega_i/\pi$) or weights for 10 different downwelling radiances when representing the reflected component.
The angles represent zenith angles, with $0^{\circ}$ corresponding to overhead incidence and $90^{\circ}$ to grazing incidence, parallel to the ground.
The most reflective parts of the scene---the checkerboards---are placed almost vertically oriented, resulting in near grazing incidence due to their surface normals being roughly horizontal.
The weights of downwelling radiance at angles $70^{\circ}$ and $80^{\circ}$ are the highest, aligning with the orientation of the checkerboard targets.
This is very encouraging, as accounting for downwelling radiance not only corrects the artifacts but also can provide additional information such as the orientation given that downwelling radiance changes significantly depending on the zenith angle. 
As the emissivity of the object grows, the estimation of weights is less reasonable because the contribution from the reflected radiance becomes negligible.

\begin{figure}
    \centering
     \begin{subfigure}[t]{0.49\linewidth}
        \centering
        \includegraphics[width=\linewidth]{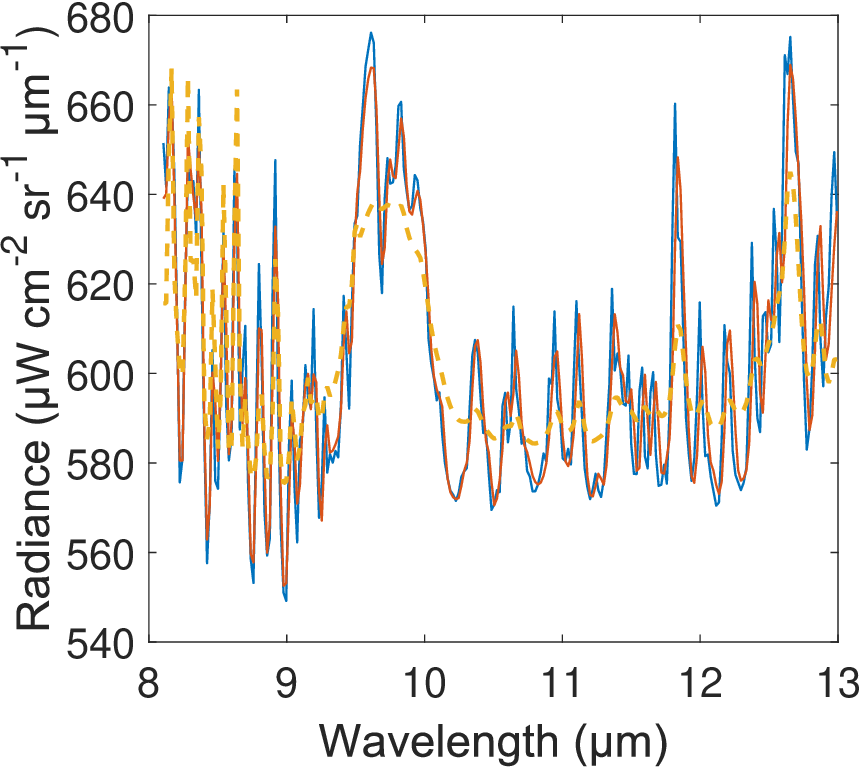}
        \caption{Reflective panel pixel (reflective part)}
        \label{fig:Experimental_model_fit1}
    \end{subfigure}
    \hfill
    \centering
    \begin{subfigure}[t]{0.49\linewidth}
        \centering
        \includegraphics[width=\linewidth]{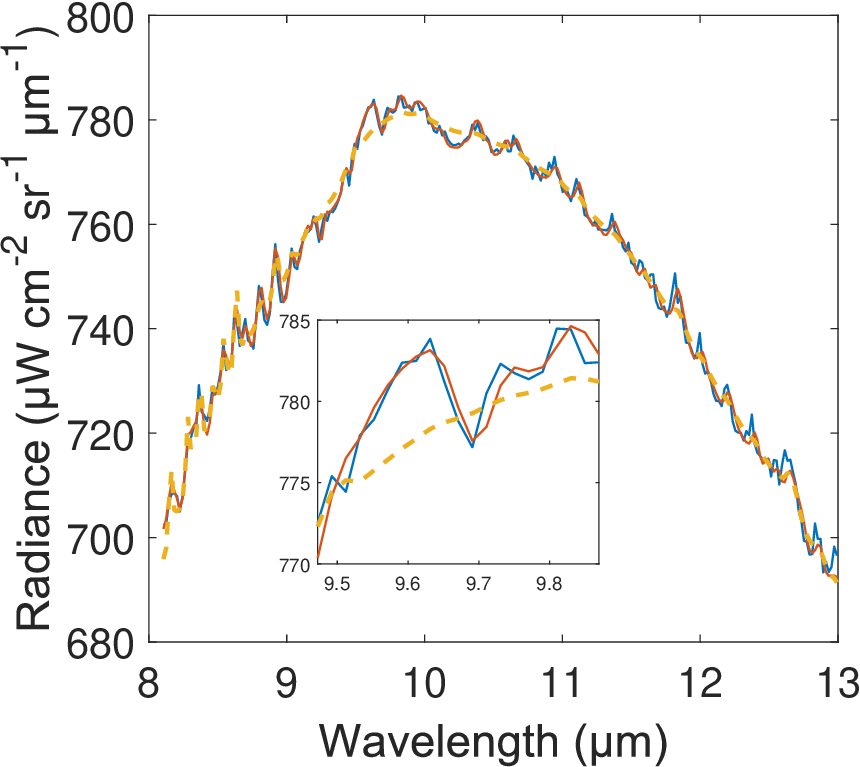}
        \caption{Reflective panel pixel (less reflective part)}
        \label{fig:Experimental_model_fit2}
    \end{subfigure}
    \centering
    \hfill
    \begin{subfigure}[t]{0.49\linewidth}
        \centering
        \includegraphics[width=\linewidth]{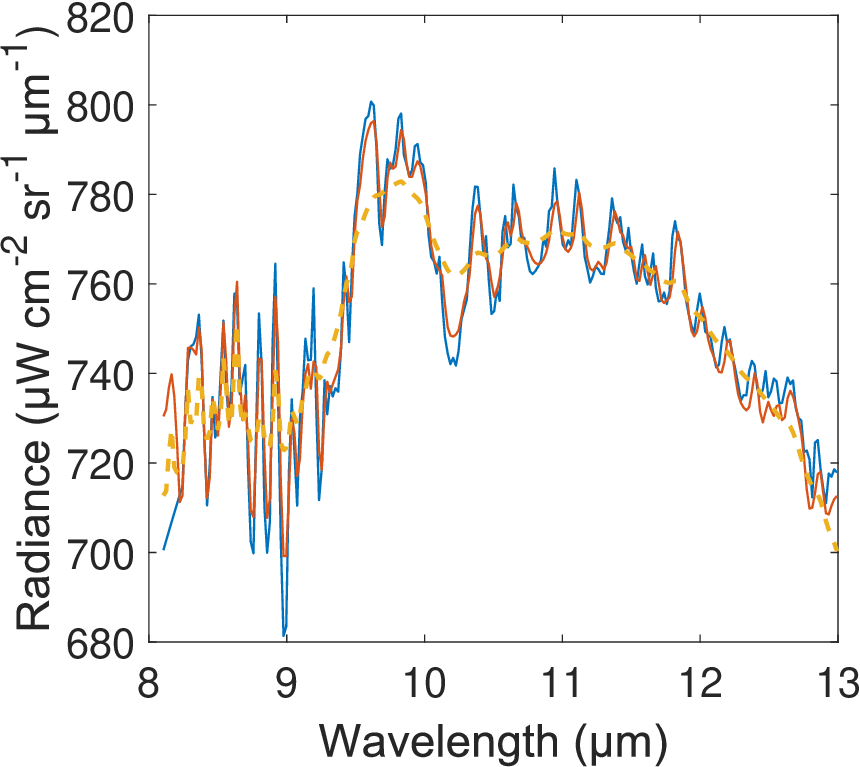}
        \caption{Sky pixel}
        \label{fig:Experimental_model_fit3}
    \end{subfigure}
    \begin{subfigure}[t]{0.49\linewidth}
        \centering
        \includegraphics[width=\linewidth]{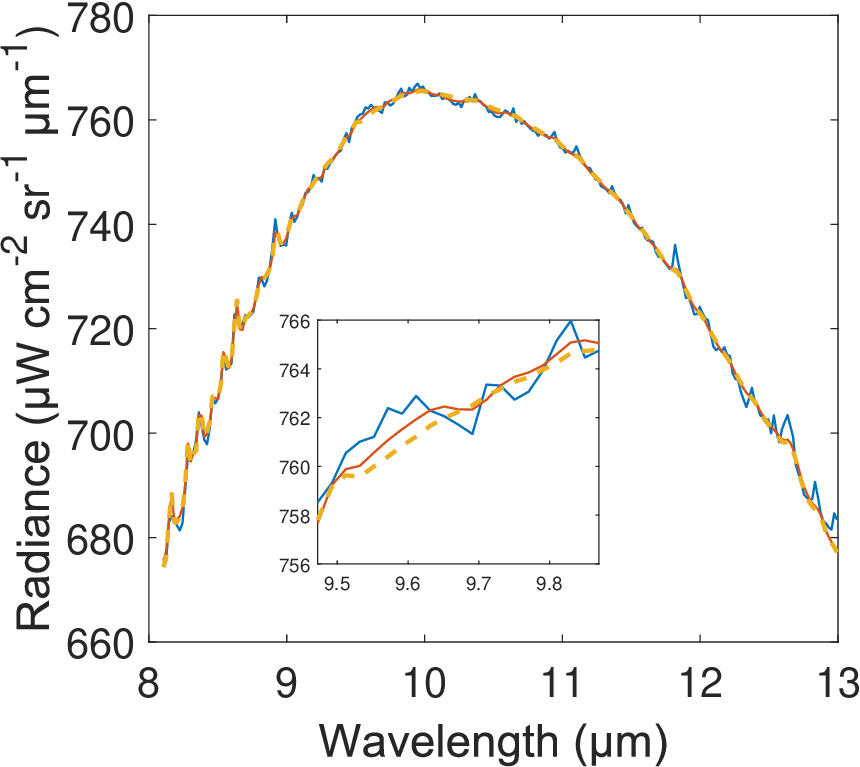}
        \caption{Grass pixel}
        \label{fig:Experimental_model_fit4}
    \end{subfigure}
    \centering
    \hfill
    \begin{subfigure}[t]{0.49\linewidth}
        \centering
        \includegraphics[width=\linewidth]{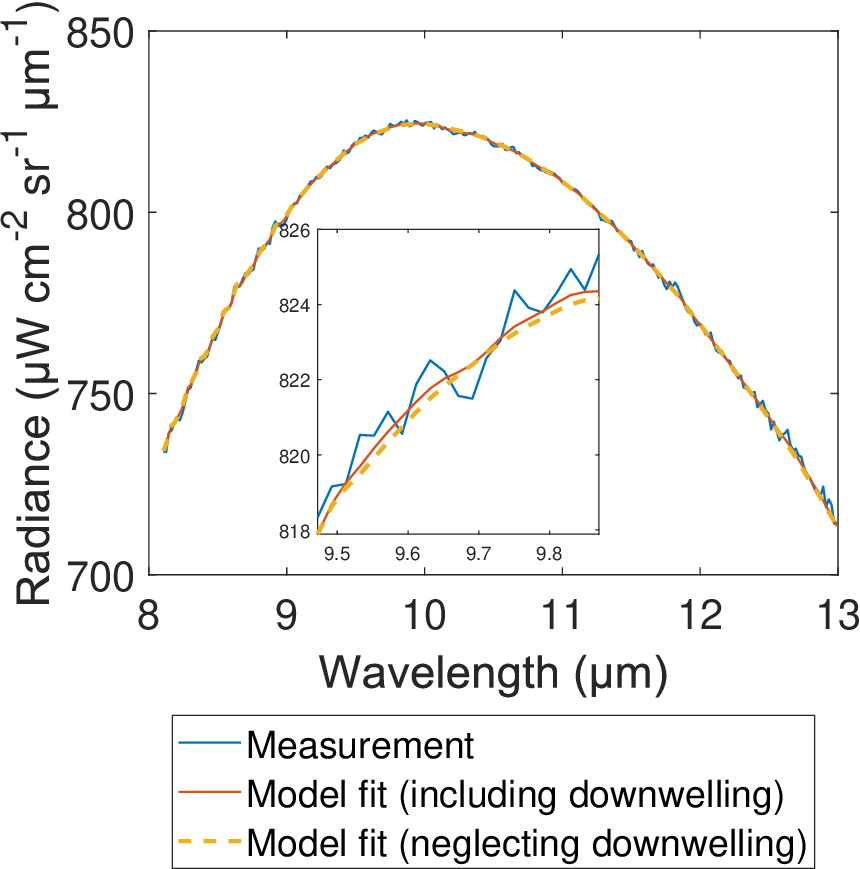}
        \caption{Foreground tree pixel}
        \label{fig:Experimental_model_fit5}
    \end{subfigure}
    \hfill
    \begin{subfigure}[t]{0.49\linewidth}
        \centering
        \includegraphics[width=\linewidth]{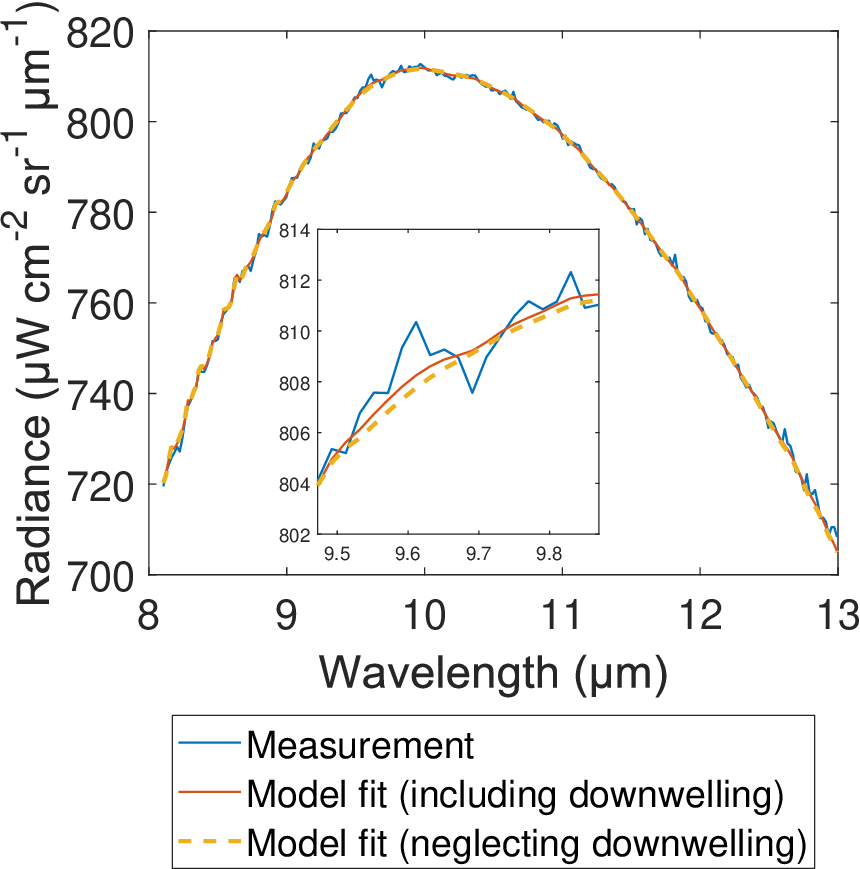}
        \caption{Background forest pixel}
        \label{fig:Experimental_model_fit6}
    \end{subfigure}
    \caption{Model fits for 6 pixels accross the scene, \figref{fig:Experimental_model_fit1} more reflective part of the checkerboard target, \figref{fig:Experimental_model_fit2} less reflective part of the checkerboard target, \figref{fig:Experimental_model_fit3} sky, \figref{fig:Experimental_model_fit4} grass, \figref{fig:Experimental_model_fit5} foreground tree, \figref{fig:Experimental_model_fit6} background forest. In each plot the solid blue curve represents the measurements, the solid orange curve represents the model fit accounting for reflected downwelling radiance and dashed yellow curve represent the model fit ignoring downwelling radiance.}
    \label{fig:Experimental_model_fit}
\end{figure}

\Cref{fig:Experimental_model_fit} presents the model fits for six representative pixels across the scene.
In each plot, the solid blue line shows the measurements,
the solid orange curve represents the model fit including downwelling radiance,
and the dashed yellow line represents the model fit ignoring downwelling radiance.
For clarity, we zoom-in on the ozone absorption line, where the effect of downwelling radiance can be seen clearly.
\Cref{fig:Experimental_model_fit1,fig:Experimental_model_fit2} show results for the checkerboard target's more reflective and less reflective areas, respectively.
The more reflective area displays noticeably stronger absorption features, as it reflects more downwelling radiance.
In both cases, incorporating downwelling radiance into the model leads to better fit and more accurate estimates.
\Cref{fig:Experimental_model_fit3} corresponds to a sky pixel at the top of the image, where accounting for downwelling radiance similarly improves the fit.
\Cref{fig:Experimental_model_fit4,fig:Experimental_model_fit5,fig:Experimental_model_fit6}
correspond to highly emissive pixels in the scene.
In these cases, downwelling radiance has minimal influence due to the low level of reflection, and as expected, the model fits nearly identically whether downwelling radiance is considered or not.

To further illustrate the hyperspectral ranging method, \cref{fig:Experimental_hyperspectral_results_other_results} shows
results for five additional scenes
from dataset IHTest\_202104
for which the necessary atmospheric measurements were available.
The atmospheric parameters used in the estimation---air temperature, relative humidity, and pressure---are given in each column.
Sky pixels are identified based on the estimated downwelling contribution exceeding a fixed threshold; these are assigned the maximum range value in the visualization.

We see qualitative improvements from the mitigation of downwelling radiance introduced here.
In the first three columns,
distance estimates for the reflective calibration targets are improved, with remaining errors
possibly due
to model mismatch arising from the assumption of a cloudless standard atmosphere in the downwelling radiance calculation, as well as atmospheric inhomogeneities.
Improvements are also observed over natural scene elements, particularly tree trunks and forested regions in the second and third columns.
In the third column, increased inaccuracies in the nearby grass region
may be due
to the small temperature contrast between the grass and the ambient air, which makes the estimation more sensitive to uncertainties in the recorded atmospheric parameters and local air inhomogeneities.
Note that the higher humidity conditions
(first three columns)
are favorable for absorption-based ranging, given the spectral range of the instrument.
Since thermal imaging instruments typically operate in non-absorptive spectral bands, higher humidity enables more effective use of water vapor absorption in the 8–9~\si{\micro\meter} range.
This results in cleaner distance estimates in Paths~2 and~15 (first three columns) compared to Path~26 (last two columns).
We note that an instrument operating directly within absorptive spectral bands would further improve ranging accuracy, as discussed in \cite{gallastegi2024absorption}.

\newlength{\figHeight}
\setlength{\figHeight}{4.2cm}

\begin{figure*}
  \centering
  \begin{tabular}{@{}c@{\,\,}c@{\,\,}c@{\,\,}c@{\,\,}c@{\,\,}c@{}}
    & {\small Path 2, Step 5}
    & {\small Path 15, Step 4}
    & {\small Path 15, Step 8}
    & {\small Path 26, Step 1}
    & {\small Path 26, Step 5}
   \\
    \rotatebox[origin=l]{90}{\small \qquad RGB image}
    & \includegraphics[height=0.74\figHeight]{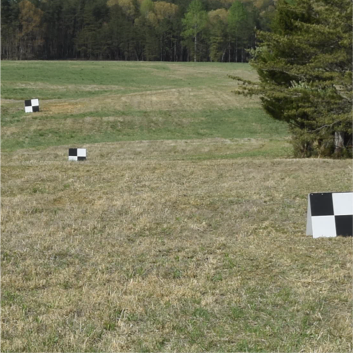}
    & \includegraphics[height=0.74\figHeight]{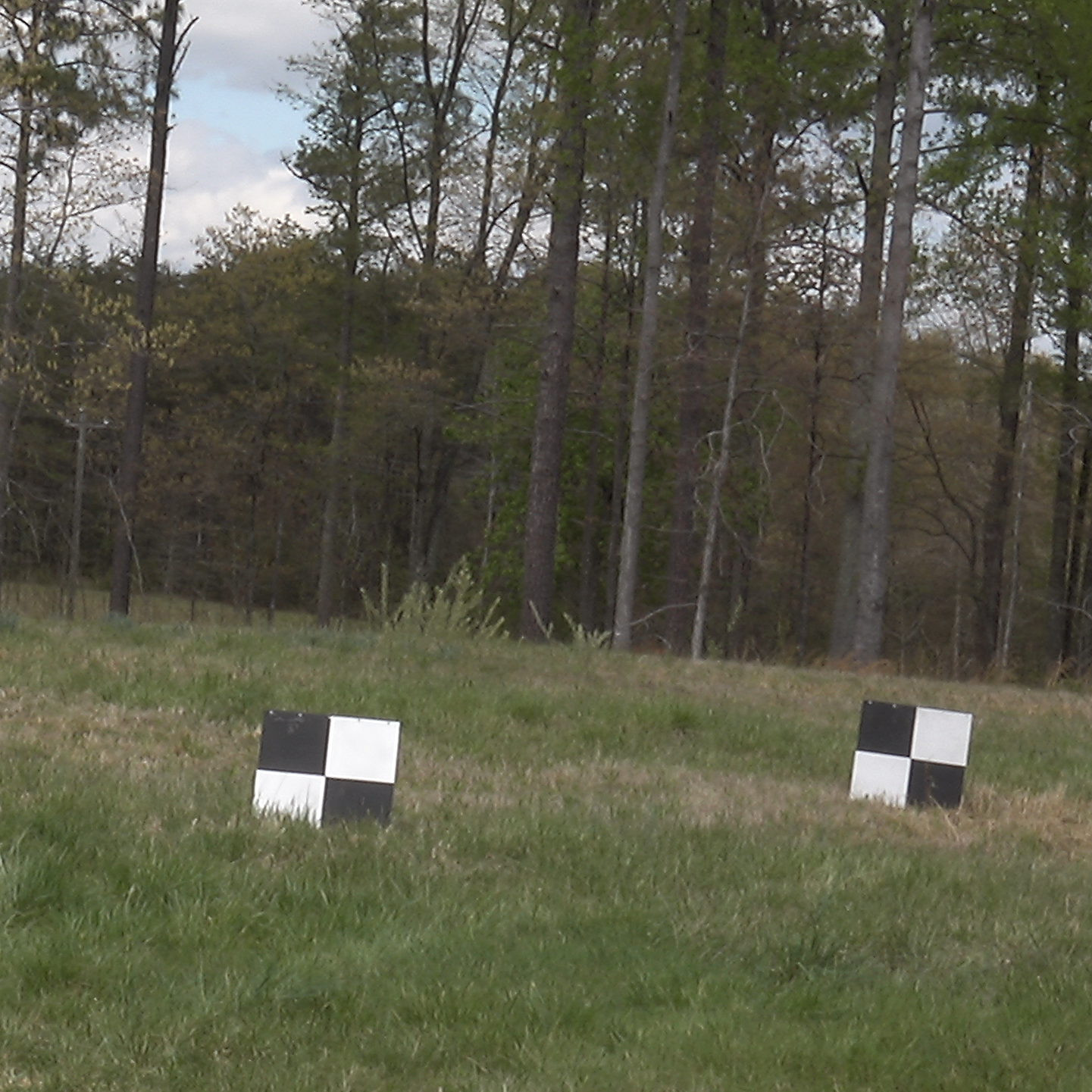}
    & \includegraphics[height=0.74\figHeight]{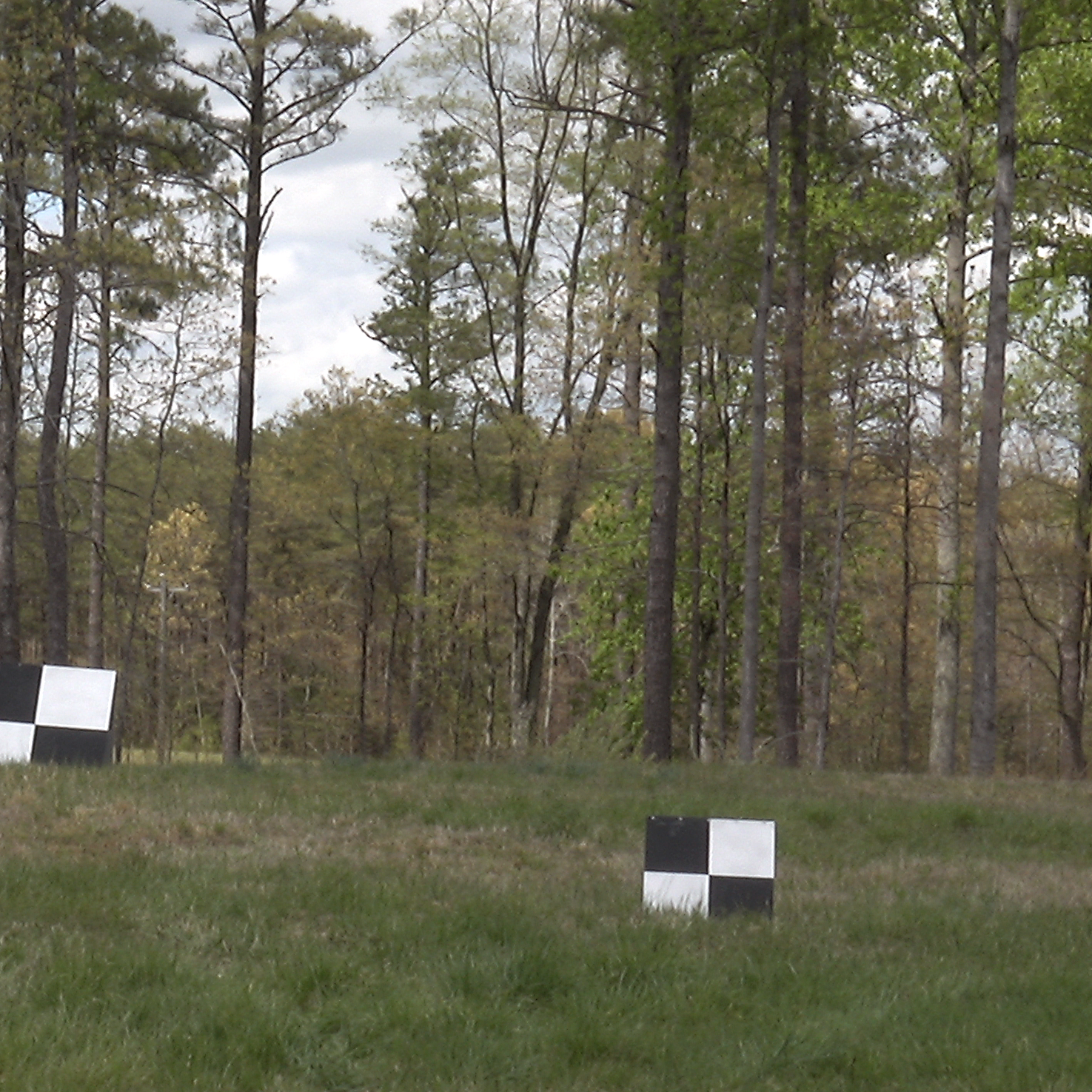}
    & \includegraphics[height=0.74\figHeight]{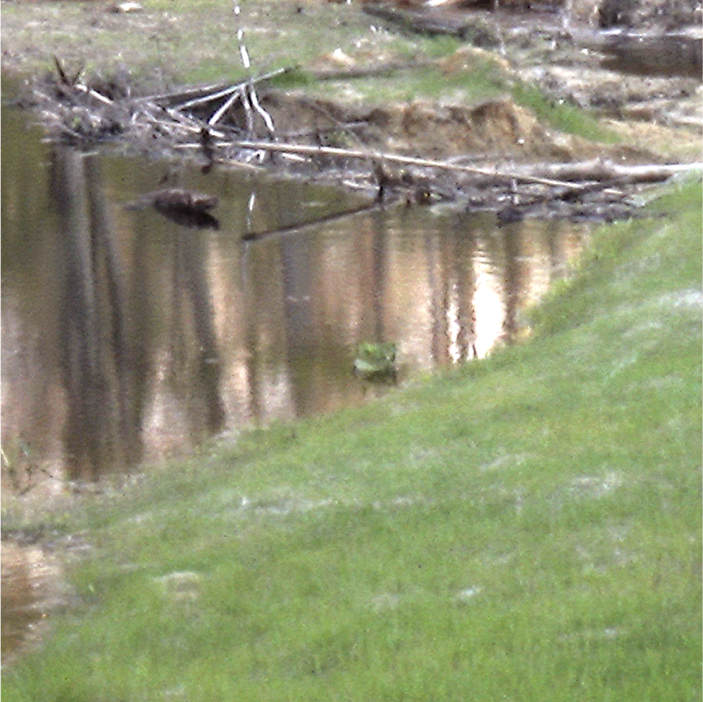}
    & \includegraphics[height=0.74\figHeight]{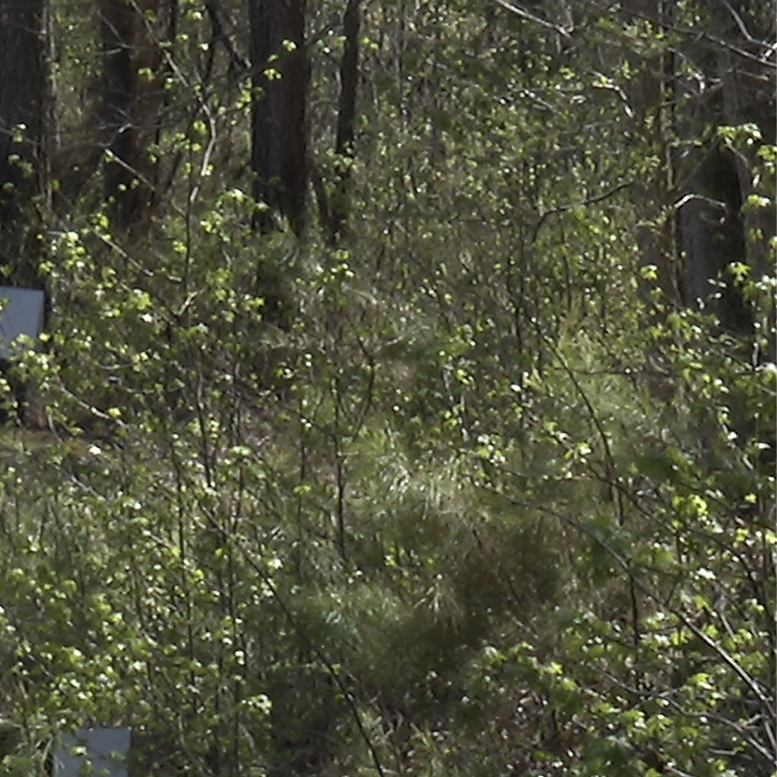}
   \\
    \rotatebox[origin=l]{90}{\small Neglecting downwelling}
    & \includegraphics[height=\figHeight]{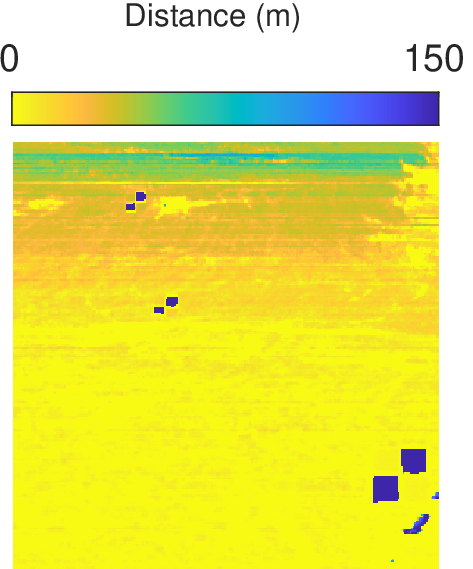}
    & \includegraphics[height=\figHeight]{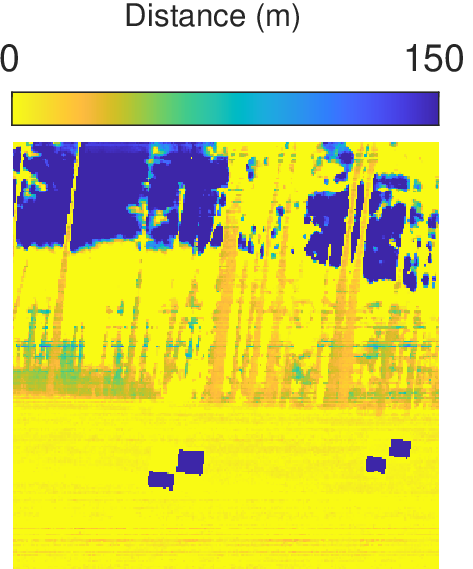}
    & \includegraphics[height=\figHeight]{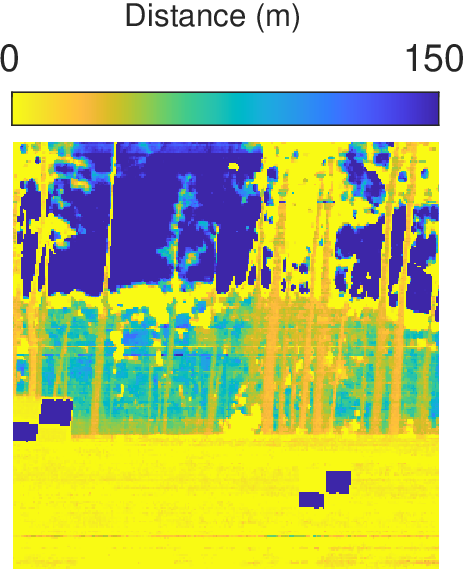}
    & \includegraphics[height=\figHeight]{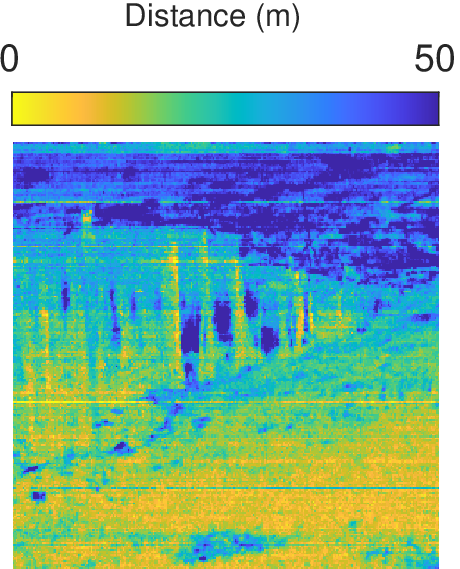}
    & \includegraphics[height=\figHeight]{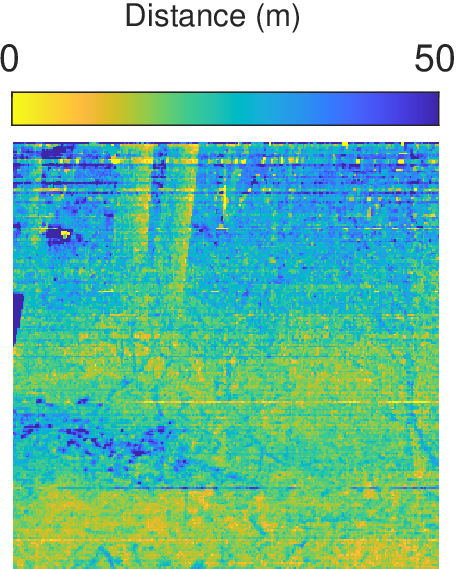}
   \\
    \rotatebox[origin=l]{90}{\small Including downwelling}
    & \includegraphics[height=\figHeight]{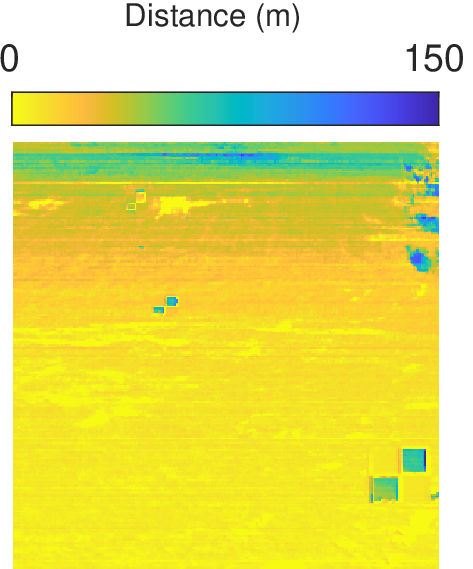}
    & \includegraphics[height=\figHeight]{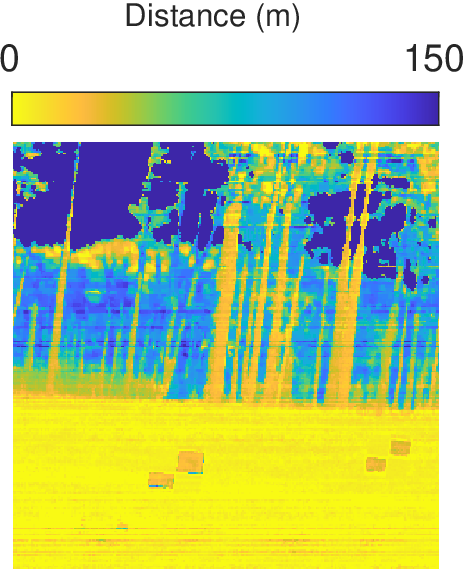}
    & \includegraphics[height=\figHeight]{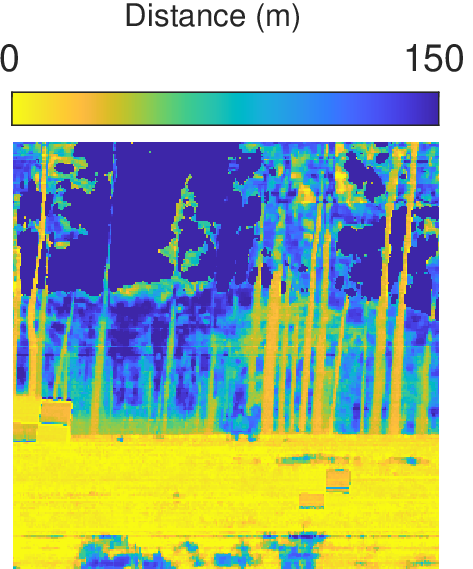}
    & \includegraphics[height=\figHeight]{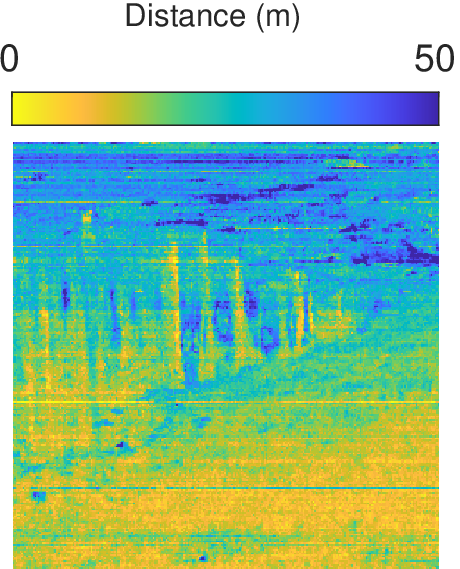}
    & \includegraphics[height=\figHeight]{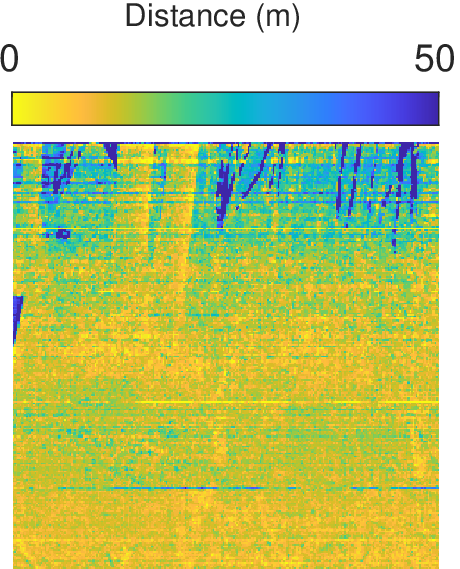}
   \\
    & {\scriptsize $\Tair = 291.88\,\si{K}$}
    & {\scriptsize $\Tair = 289.70\,\si{K}$}
    & {\scriptsize $\Tair = 289.70\,\si{K}$}
    & {\scriptsize $\Tair = 294.88\,\si{K}$}
    & {\scriptsize $\Tair = 294.61\,\si{K}$}
   \\
    & {\scriptsize Relative humidity: 53.28\%}
    & {\scriptsize Relative humidity: 63.39\%}
    & {\scriptsize Relative humidity: 63.39\%}
    & {\scriptsize Relative humidity: 31.57\%}
    & {\scriptsize Relative humidity: 31.71\%}
   \\
    & {\scriptsize Pressure: 1009.13\,\si{\milli\bar}}
    & {\scriptsize Pressure: 1009.31\,\si{\milli\bar}}
    & {\scriptsize Pressure: 1009.31\,\si{\milli\bar}}
    & {\scriptsize Pressure: 1003.49\,\si{\milli\bar}}
    & {\scriptsize Pressure: 1003.51\,\si{\milli\bar}}
  \end{tabular}
  \caption{Hyperspectral absorption-based ranging results on additional scenes.
    Each column corresponds to a specific path and step from the dataset available in~\cite{Yellin_2024_WACV}.
    The first row shows the RGB images for visual reference.
    Results in the second row are obtained with downwelling neglected~\cite{10877411},
    and results in the third row use the downwelling mitigation introduced here.
    }
    \label{fig:Experimental_hyperspectral_results_other_results}
\end{figure*}

\section{Conclusion}

In this work, we addressed the challenge of artifacts caused by downwelling radiance in passive absorption-based ranging.
The effect of downwelling radiance is especially prominent for natural scenes where temperature variations are low and the contribution from reflected environmental thermal sources cannot be ignored.
We leverage the ozone cues in the downwelling radiance to mitigate the errors.
We presented two methods---a quadspectral approach and a hyperspectral approach---that effectively mitigate these errors by incorporating downwelling radiance into the model.
The quadspectral approach leverages strong cues on ozone absorption from two wavelengths to correct range estimates.
The hyperspectral approach uses a number of precomputed downwelling radiances,
providing richer information with fewer assumptions,
making it a promising tool for passive ranging in complex natural environments.
By utilizing a larger portion of the infrared spectrum, the accuracy improves and enables the estimation of both temperature and emissivity profiles for each pixel.
Our experimental results demonstrate that accounting for downwelling radiance significantly improves the ranging accuracy across the scene, particularly in highly reflective regions. 
For example, for the front checkerboard target in our test scene, errors relative to lidar are reduced from over 100\,\si{\meter} when downwelling is unmodeled to about
6.8\,\si{\meter} for the quadspectral method
(see \cref{tab:Experimental_bispectral_table})
and
1.2\,\si{\meter} for the hyperspectral method
(see \cref{tab:Experimental_hyperspectral_table}).
Accounting for downwelling radiance at various zenith angles corrects the artifacts in temperature and emissivity estimates with hyperspectral measurements and can provide object orientation information for highly reflective materials.
We note that the method’s performance depends on atmospheric parameters such as temperature, humidity, and pressure. Higher humidity generally improves performance in the LWIR band, and larger temperature contrasts between objects and the air increase sensitivity to distance. More details on how atmospheric parameters affect ranging performance can be found in~\cite{gallastegi2024absorption}.

\section*{Code Availability}
The code for the proposed method is publicly available at
\href{https://github.com/unaydorken/Ozone-Cues-Mitigate-Reflected-Downwelling-Radiance-in-LWIR-Absorption-Based-Ranging}{https://github.com/unaydorken/Ozone-Cues-Mitigate-Reflected-Downwelling-Radiance-in-LWIR-Absorption-Based-Ranging}.

\section*{Acknowledgment}
Data provided by the U.S. Army Night Vision and Electronic Sensors Directorate 
and Johns Hopkins University Applied Physics Laboratory.

\newcommand{\SortNoop}[1]{}


\begin{thebibliography}{10}
\providecommand{\url}[1]{#1}
\csname url@samestyle\endcsname
\providecommand{\newblock}{\relax}
\providecommand{\bibinfo}[2]{#2}
\providecommand{\BIBentrySTDinterwordspacing}{\spaceskip=0pt\relax}
\providecommand{\BIBentryALTinterwordstretchfactor}{4}
\providecommand{\BIBentryALTinterwordspacing}{\spaceskip=\fontdimen2\font plus
\BIBentryALTinterwordstretchfactor\fontdimen3\font minus \fontdimen4\font\relax}
\providecommand{\BIBforeignlanguage}[2]{{%
\expandafter\ifx\csname l@#1\endcsname\relax
\typeout{** WARNING: IEEEtran.bst: No hyphenation pattern has been}%
\typeout{** loaded for the language `#1'. Using the pattern for}%
\typeout{** the default language instead.}%
\else
\language=\csname l@#1\endcsname
\fi
#2}}
\providecommand{\BIBdecl}{\relax}
\BIBdecl

\bibitem{LiuPRH:2020}
Y.~Liu, N.~Pears, P.~L. Rosin, and P.~Huber, Eds., \emph{3D Imaging, Analysis and Applications}, 2nd~ed.\hskip 1em plus 0.5em minus 0.4em\relax Springer, 2020.

\bibitem{matthies1994stochastic}
L.~Matthies and P.~Grandjean, ``Stochastic performance, modeling and evaluation of obstacle detectability with imaging range sensors,'' \emph{IEEE Trans. Robot. Automat.}, vol.~10, no.~6, pp. 783--792, Dec. 1994.

\bibitem{sibley2007bias}
G.~Sibley, L.~Matthies, and G.~Sukhatme, ``Bias reduction and filter convergence for long range stereo,'' in \emph{Robotics Research}, ser. Springer Tracts in Advanced Robotics, S.~Thrun, R.~Brooks, and H.~Durrant-Whyte, Eds.\hskip 1em plus 0.5em minus 0.4em\relax Springer, 2007, vol.~28, pp. 285--294.

\bibitem{Clark2010}
D.~E. Clark and \v{S}.~Ivekovi\v{c}, ``The {C}ramer-{R}ao lower bound for {3-D} state estimation from rectified stereo cameras,'' in \emph{13th Int. Conf. Information Fusion}, 2010, pp. 1--8.

\bibitem{Gao2020}
B.~Gao, H.~Lang, and J.~Ren, ``Stereo visual {SLAM} for autonomous vehicles: A review,'' in \emph{IEEE Int. Conf. Systems, Man, and Cybernetics (SMC)}, 2020, pp. 1316--1322.

\bibitem{Gurton:14}
K.~P. Gurton, A.~J. Yuffa, and G.~W. Videen, ``Enhanced facial recognition for thermal imagery using polarimetric imaging,'' \emph{Opt. Lett.}, vol.~39, no.~13, pp. 3857--3859, Jul. 2014.

\bibitem{bao2023heat}
F.~Bao, X.~Wang, S.~H. Sureshbabu, G.~Sreekumar, L.~Yang, V.~Aggarwal, V.~N. Boddeti, and Z.~Jacob, ``Heat-assisted detection and ranging,'' \emph{Nature}, vol. 619, no. 7971, pp. 743--748, 2023.

\bibitem{bao2024thermal}
F.~Bao, S.~Jape, A.~Schramka, J.~Wang, T.~E. McGraw, and Z.~Jacob, ``Why thermal images are blurry,'' \emph{Opt. Express}, vol.~32, no.~3, pp. 3852--3865, 2024.

\bibitem{leonpacher1983passive}
N.~K. Leonpacher, ``Passive infrared ranging,'' Master's thesis, Air Force Institute of Technology, Dec. 1983.

\bibitem{hawks2006passive}
M.~R. Hawks, ``Passive ranging using atmospheric oxygen absorption spectra,'' Ph.D. dissertation, Air Force Institute of Technology, Mar. 2006.

\bibitem{anderson2010monocular}
J.~R. Anderson, ``Monocular passive ranging by an optical system with band pass filtering,'' Master's thesis, Air Force Institute of Technology, Mar. 2010.

\bibitem{anderson2011flight}
J.~R. Anderson, M.~R. Hawks, K.~C. Gross, and G.~P. Perram, ``Flight test of an imaging {O$_2$(X-b)} monocular passive ranging instrument,'' in \emph{Proc. SPIE Airborne Intelligence, Surveillance, Reconnaissance (ISR) Systems and Applications VIII}, vol. 8020, 2011, pp. 38--49.

\bibitem{vincent2011passive}
R.~A. Vincent and M.~R. Hawks, ``Passive ranging of dynamic rocket plumes using infrared and visible oxygen attenuation,'' in \emph{Proc. SPIE Acquisition, Tracking, Pointing, and Laser Systems Technologies XXV}, vol. 8052, 2011, p. 80520D.

\bibitem{hawks2013short}
M.~R. Hawks, R.~A. Vincent, J.~Martin, and G.~P. Perram, ``Short-range demonstrations of monocular passive ranging using {O$_2$ (X$^3\Sigma_g^- \rightarrow b^1\Sigma^+_g$}) absorption spectra,'' \emph{Appl. Spectroscopy}, vol.~67, no.~5, pp. 513--519, 2013.

\bibitem{yu2017passive}
H.~Yu, B.~Liu, Z.~Yan, and Y.~Zhang, ``Passive ranging using a filter-based non-imaging method based on oxygen absorption,'' \emph{Applied Optics}, vol.~56, no.~28, pp. 7803--7807, 2017.

\bibitem{yu2019threechannel}
H.~Yu, B.~Liu, Y.~Zhang, Z.~Yan, Y.~Chen, S.~Zhang, F.~Huang, and X.~Shen, ``Three-channel filter-based non-imaging passive ranging system based on oxygen absorption,'' \emph{Optics \& Laser Technology}, vol. 111, pp. 797--801, Apr. 2019.

\bibitem{Nagase2022}
Y.~Nagase, T.~Kushida, K.~Tanaka, T.~Funatomi, and Y.~Mukaigawa, ``Shape from thermal radiation: Passive ranging using multi-spectral {LWIR} measurements,'' in \emph{Proc. IEEE Conf. Comput. Vis. Pattern Recog.}, New Orleans, LA, Jun. 2022, pp. 12\,661--12\,671.

\bibitem{10877411}
U.~Dorken~Gallastegi, H.~Rueda-Chac{\'o}n, M.~J. Stevens, and V.~K. Goyal, ``Absorption-based, passive range imaging from hyperspectral thermal measurements,'' \emph{IEEE Trans. Patt. Anal. Mach. Intell.}, vol.~47, no.~5, pp. 4044--4060, May 2025.

\bibitem{Kaariainen2024}
T.~K{\"a}{\"a}ri{\"a}inen and J.~Sepp{\"a}, ``{3D} camera based on laser light absorption by atmospheric oxygen at 761 nm,'' \emph{Opt. Express}, vol.~32, no.~4, pp. 6342--6349, 2024.

\bibitem{Kushida2024}
T.~Kushida, R.~Nakamura, H.~Matsuda, W.~Chen, and K.~Tanaka, ``Affine transform representation for reducing calibration cost on absorption-based {LWIR} depth sensing,'' \emph{Sci. Rep.}, vol.~14, p. 26429, Nov. 2024.

\bibitem{cermak2017eleven}
V.~Cermak, L.~Bodri, M.~Kresl, P.~Dedecek, and J.~Safanda, ``Eleven years of ground--air temperature tracking over different land cover types,'' \emph{Int. J. Climatology}, vol.~37, no.~2, pp. 1084--1099, 2017.

\bibitem{gallastegi2024absorption}
U.~Dorken~Gallastegi, H.~Rueda-Chac{\'o}n, M.~J. Stevens, and V.~K. Goyal, ``Absorption-based hyperspectral thermal ranging: performance analyses, optimization, and simulations,'' \emph{Optics Express}, vol.~32, no.~1, pp. 151--166, 2024.

\bibitem{Higashiyama2012}
A.~Higashiyama and K.~Shimono, ``Apparent depth of pictures reflected by a mirror: The plastic effect,'' \emph{Atten. Percept. Psychophys.}, vol.~74, no.~7, pp. 1522--1532, Oct. 2012.

\bibitem{Yang2011}
S.-W. Yang and C.-C. Wang, ``On solving mirror reflection in {LIDAR} sensing,'' \emph{IEEE/ASME Trans. Mechatronics}, vol.~16, no.~2, pp. 255--265, Apr. 2011.

\bibitem{Zhao2020}
X.~Zhao, Z.~Yang, and S.~Schwertfeger, ``Mapping with reflection - detection and utilization of reflection in {3D} lidar scans,'' in \emph{Proc. IEEE Symp. Safety, Security, and Rescue Robotics}, Abu Dhabi, UAE, Nov. 2020, pp. 27--33.

\bibitem{FaccioVW:20}
D.~Faccio, A.~Velten, and G.~Wetzstein, ``Non-line-of-sight imaging,'' \emph{Nat. Rev. Phys.}, vol.~2, no.~6, pp. 318--327, Jun. 2020.

\bibitem{BhatN:98}
D.~N. Bhat and S.~K. Nayar, ``Stereo and specular reflection,'' \emph{Int. J. Comput. Vis.}, vol.~26, no.~2, pp. 91--106, Feb. 1998.

\bibitem{Whelan2018}
T.~Whelan, M.~Goesele, S.~J. Lovegrove, J.~Straub, S.~Green, R.~Szeliski, S.~Butterfield, S.~Verma, and R.~Newcombe, ``Reconstructing scenes with mirror and glass surfaces,'' \emph{ACM Trans. Graph.}, vol.~37, no.~4, Aug. 2018, article 102.

\bibitem{Manolakis2019}
D.~Manolakis, M.~Pieper, E.~Truslow, R.~Lockwood, A.~Weisner, J.~Jacobson, and T.~Cooley, ``Longwave infrared hyperspectral imaging: Principles, progress, and challenges,'' \emph{IEEE Geosci. Remote Sensing Mag.}, vol.~7, no.~2, pp. 72--100, Jun. 2019.

\bibitem{borel2008error}
C.~Borel, ``Error analysis for a temperature and emissivity retrieval algorithm for hyperspectral imaging data,'' \emph{Int. J. Remote Sensing}, vol.~29, no. 17--18, pp. 5029--5045, Sep. 2008.

\bibitem{borel2011recent}
C.~C. Borel and R.~F. Tuttle, ``Recent advances in temperature-emissivity separation algorithms,'' in \emph{2011 Aerospace Conference}.\hskip 1em plus 0.5em minus 0.4em\relax IEEE, 2011, pp. 1--14.

\bibitem{10.1117/12.2239138}
M.~Pieper, D.~Manolakis, E.~Truslow, T.~Cooley, M.~Brueggeman, A.~Weisner, and J.~Jacobson, ``{In-scene LWIR downwelling radiance estimation},'' in \emph{SPIE Imaging Spectrometry XXI}, J.~F. Silny and E.~J. Ientilucci, Eds., vol. 9976, 2016, p. 99760E.

\bibitem{adler2014long}
S.~Adler-Golden, P.~Conforti, M.~Gagnon, P.~Tremblay, and M.~Chamberland, ``Long-wave infrared surface reflectance spectra retrieved from telops hyper-cam imagery,'' in \emph{SPIE Algorithms and Technologies for Multispectral, Hyperspectral, and Ultraspectral Imagery XX}, vol. 9088, 2014, pp. 247--254.

\bibitem{asano2020depth}
Y.~Asano, Y.~Zheng, K.~Nishino, and I.~Sato, ``Depth sensing by near-infrared light absorption in water,'' \emph{IEEE Trans. Patt. Anal. Mach. Intell.}, vol.~43, no.~8, pp. 2611--2622, 2020.

\bibitem{kuo2021non}
M.-Y.~J. Kuo, R.~Kawahara, S.~Nobuhara, and K.~Nishino, ``Non-rigid shape from water,'' \emph{IEEE Trans. Patt. Anal. Mach. Intell.}, vol.~43, no.~7, pp. 2220--2232, Jul. 2021.

\bibitem{kuo2021surface}
M.~J. Kuo, S.~Murai, R.~Kawahara, S.~Nobuhara, and K.~Nishino, ``Surface normals and shape from water,'' \emph{IEEE Trans. Patt. Anal. Mach. Intell.}, vol.~44, no.~12, pp. 9150--9162, Dec. 2022.

\bibitem{tsiotsios2017near}
C.~Tsiotsios, A.~J. Davison, and T.-K. Kim, ``Near-lighting photometric stereo for unknown scene distance and medium attenuation,'' \emph{Image and Vision Computing}, vol.~57, pp. 44--57, 2017.

\bibitem{fujimura2018photometric}
Y.~Fujimura, M.~Iiyama, A.~Hashimoto, and M.~Minoh, ``Photometric stereo in participating media considering shape-dependent forward scatter,'' in \emph{Proc. IEEE Conf. Comput. Vis. Pattern Recog.}, 2018, pp. 7445--7453.

\bibitem{gallastegi2022absorption}
U.~Dorken~Gallastegi, H.~Rueda-Chacon, M.~J. Stevens, and V.~K. Goyal, ``Absorption-based ranging from ambient thermal radiation without known emissivities,'' in \emph{Optica Conf. Lasers Electro-Optics}, 2022, presentation STh5J.3 online.

\bibitem{minnaert1995light}
M.~Minnaert, \emph{Light and Color in the Outdoors}.\hskip 1em plus 0.5em minus 0.4em\relax Springer Science \& Business Media, 1995, vol.~17.

\bibitem{manolakis2016hyperspectral}
D.~G. Manolakis, R.~B. Lockwood, and T.~W. Cooley, \emph{Hyperspectral Imaging Remote Sensing: Physics, Sensors, and Algorithms}.\hskip 1em plus 0.5em minus 0.4em\relax Cambridge University Press, 2016.

\bibitem{dubuc2021design}
A.~Dubuc, ``Design of a reflective infrared spectrograph for exoplanet spectroscopy,'' Master's thesis, Universit{\'e} de Li{\`e}ge, Li{\`e}ge, Belgique, 2021.

\bibitem{SpectralCalc}
``SpectralCalc: High-resolution spectral modeling,'' \url{https://www.spectralcalc.com}, accessed: 2022-06-05.

\bibitem{GORDON2022107949}
I. E. Gordon \emph{et al.},
``The {HITRAN2020} molecular spectroscopic database,'' \emph{J. Quantitative Spectroscopy and Radiative Transfer}, vol. 277, p. 107949, 2022.

\bibitem{beirle2017parameterizing}
S.~Beirle, J.~Lampel, C.~Lerot, H.~Sihler, and T.~Wagner, ``Parameterizing the instrumental spectral response function and its changes by a super-{G}aussian and its derivatives,'' \emph{Atmospheric Measurement Techniques}, vol.~10, no.~2, pp. 581--598, 2017.

\bibitem{coesa1976standard}
{National Oceanic and Atmospheric Administration, National Aeronautics and Space Adminsistration, and United States Air Force}, \emph{U.S. Standard Atmosphere, 1976}.\hskip 1em plus 0.5em minus 0.4em\relax U.S. Government Printing Office, 1976.

\bibitem{borel2003artemiss}
C.~C. Borel, ``{ARTEMISS} -- an algorithm to retrieve temperature and emissivity from hyper-spectral thermal image data,'' in \emph{28th Ann. GOMACTech Conf., Hyperspectral Imaging Session}, 2003.

\bibitem{MEERDINK2019111196}
S.~K. Meerdink, S.~J. Hook, D.~A. Roberts, and E.~A. Abbott, ``The {ECOSTRESS} spectral library version 1.0,'' \emph{Remote Sensing of Environment}, vol. 230, p. 111196, 2019.

\bibitem{BALDRIDGE2009711}
A.~Baldridge, S.~Hook, C.~Grove, and G.~Rivera, ``The {ASTER} spectral library version 2.0,'' \emph{Remote Sensing of Environment}, vol. 113, no.~4, pp. 711--715, 2009.

\bibitem{Arecchi2007FieldGuide}
A.~V. Arecchi, T.~Messadi, and R.~J. Koshel, \emph{Field Guide to Illumination}.\hskip 1em plus 0.5em minus 0.4em\relax Bellingham, WA: SPIE Press, 2007.

\bibitem{Yellin_2024_WACV}
F.~Yellin, S.~McCloskey, C.~Hill, E.~Smith, and B.~Clipp, ``Concurrent band selection and traversability estimation from long-wave hyperspectral imagery in off-road settings,'' in \emph{Proc. IEEE/CVF Winter Conf. Applications of Computer Vision (WACV)}, Jan. 2024.

\bibitem{incropera1996fundamentals}
F.~P. Incropera, D.~P. DeWitt, T.~L. Bergman, A.~S. Lavine \emph{et~al.}, \emph{Fundamentals of Heat and Mass Transfer}.\hskip 1em plus 0.5em minus 0.4em\relax Wiley New York, 1996, vol.~6.

\bibitem{macqueen1967some}
J.~MacQueen, ``Some methods for classification and analysis of multivariate observations,'' in \emph{Proc. Fifth Berkeley Symp. Mathematical Statistics and Probability, Vol. 1: Statistics}, vol.~5.\hskip 1em plus 0.5em minus 0.4em\relax Univ. California Press, 1967, pp. 281--298.

\end{thebibliography}
\end{document}